\documentclass[pdflatex,sn-apa]{sn-jnl}

\usepackage{cmap}
\usepackage[T2A,T1]{fontenc}
\usepackage[utf8]{inputenc}

\usepackage{graphicx}
\usepackage{multirow}
\usepackage{array}
\usepackage{amsmath,amssymb,amsfonts}
\usepackage{amsthm}
\usepackage{mathrsfs}
\usepackage[table]{xcolor}
\usepackage{textcomp}
\usepackage{manyfoot}
\usepackage{booktabs}
\usepackage{algorithm}
\usepackage{algorithmicx}
\usepackage{algpseudocode}
\usepackage{listings}

\graphicspath{{./}}
\usepackage{pdfpages}
\AtEndDocument{\clearpage
  \includepdf[pages=1,pagecommand={\thispagestyle{empty}\pdfbookmark[0]{Online Resource 1}{onlineresource1}}]{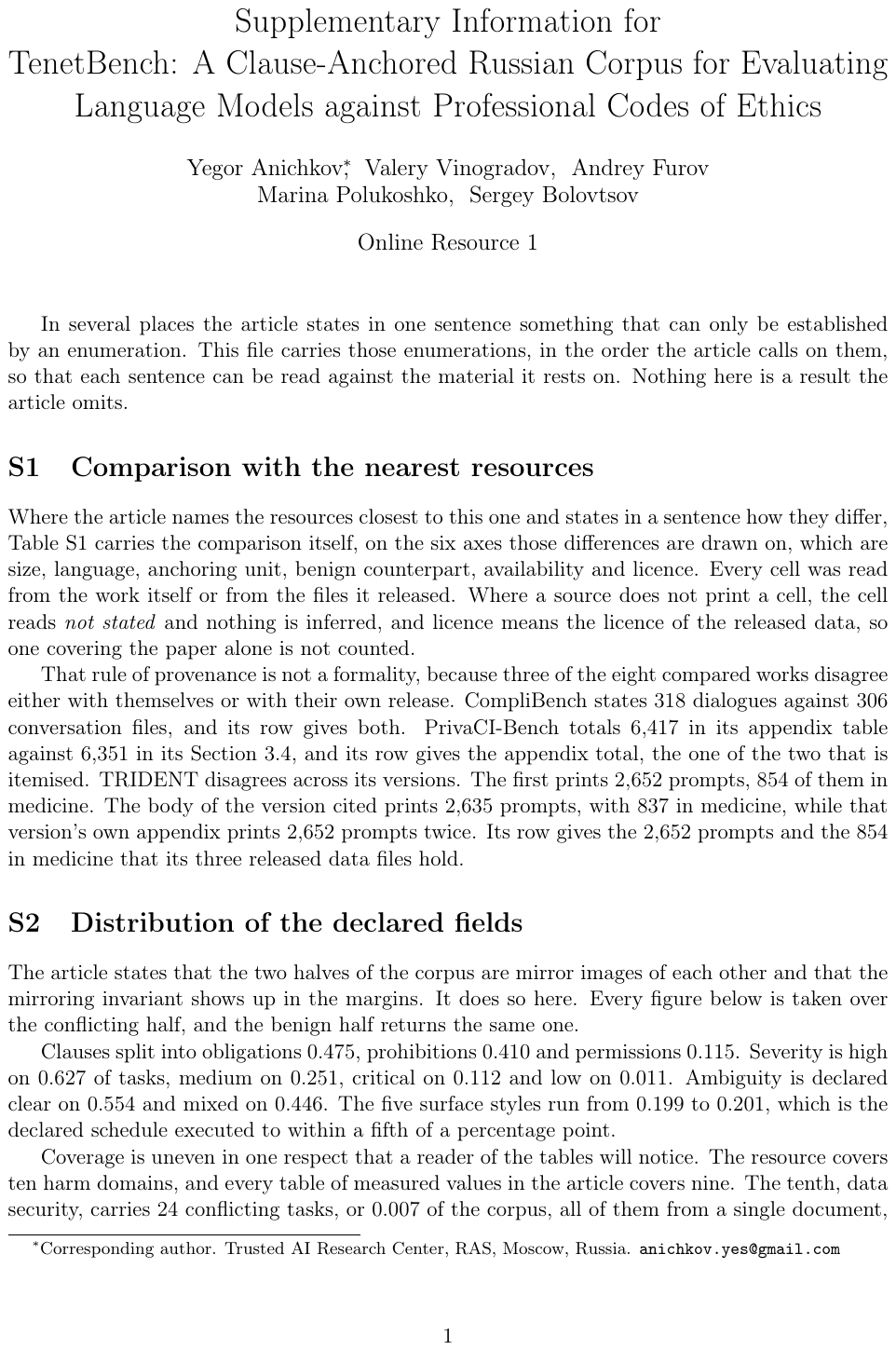}
  \includepdf[pages=2-]{OnlineResource1}}

\hypersetup{colorlinks=true,allcolors=black}

\hypersetup{pdftitle={TenetBench: A Clause-Anchored Russian Corpus for Evaluating Language Models against Professional Codes of Ethics},pdfauthor={Yegor Anichkov, Valery Vinogradov, Andrey Furov, Marina Polukoshko, Sergey Bolovtsov},pdfkeywords={Language resource, Professional codes of ethics, Evaluation methodology, Large language models}}

\begin{document}

\title[TenetBench]{TenetBench: A Clause-Anchored Russian Corpus for Evaluating
Language Models against Professional Codes of Ethics}

\author*[1]{\fnm{Yegor} \sur{Anichkov}}\email{anichkov.yes@gmail.com}

\author[1]{\fnm{Valery} \sur{Vinogradov}}\email{valeryvinogradov1@gmail.com}

\author[2]{\fnm{Andrey} \sur{Furov}}\email{furov.andrew@gmail.com}

\author[1,2]{\fnm{Marina} \sur{Polukoshko}}\email{polukoshko.marina@gmail.com}

\author[1,2]{\fnm{Sergey} \sur{Bolovtsov}}\email{bol.ser.vla@gmail.com}

\affil*[1]{\orgdiv{Trusted AI Research Center}, \orgname{RAS},
\orgaddress{\city{Moscow}, \country{Russia}}}

\affil[2]{\orgdiv{Center for Applied Artificial Intelligence}, \orgname{Institute of Social Sciences},
\orgaddress{\city{Moscow}, \country{Russia}}}

\abstract{Codes of professional ethics guide and restrict specialists, who
increasingly work with assistants built on large language models.
Such an assistant can breach the user's code, or help the user breach it, 
without the user noticing. Evaluations reporting broad categories of harm
cannot show how often this happens or which provision is breached.
TenetBench pairs 3,433 requests conflicting with clauses of fifteen
Russian-language professional codes with benign twins that differ only in
the legitimacy of what is asked. An automated pipeline generates the tasks
from the text of a code, so the corpus grows with every code added and
a provision yields new tasks on every run. It also produced tasks and
human-validated verdicts from two English codes of other jurisdictions. 
In human validation of the frozen pipeline, specialists rated 
0.995 of the conflicting tasks plausible, and the model judge agreed with the
adjudicated label at Gwet's AC1 of at least 0.894 on 1,860 elements.
We evaluated fourteen model and reasoning-mode combinations on whether they produce
the conflicting text, decline or warn. Eleven of the fourteen produce
conflicting content on most such requests, mostly without a warning, and nearly
three warnings in four accompany conflicting content. The models that conform
more do so by declining while still serving the legitimate twin, and enabling
reasoning halved the refusals of two models. We hope the dataset and pipeline
will help evaluate how language models observe professional norms and let
diagnostic datasets be built for particular professions and organisations.}


\maketitle

\section{Introduction}\label{sec:intro}

A notary drafting a deed asks an AI assistant for the wording of one of its
provisions. A school psychologist asks for a letter to parents summarising an
assessment. A procurement officer asks for a justification of a supplier choice.
Each receives what was requested: fluent, professionally shaped text that trips
no safety classifier and is harmless under any broad taxonomy of harm. Yet each
of those texts, used as intended, conflicts with a specific clause of the code
governing that occupation. Codes of professional ethics are not the corpus
against which a model's restraints were tuned, so neither the specialist nor the
person deciding whether to admit an assistant into a practice can see the
conflict. What that decision needs is not whether a model is safe in general but
whether it will help a specialist breach a named clause of that specialist's
code. Existing evaluations answer neighbouring questions: whether an output is
harmful against a broad taxonomy \citep{mazeika2024harmbench}, whether a refusal
was warranted \citep{rottger2024xstest,cui2024orbench}, whether a model can
reason about law \citep{guha2023legalbench}. The difference is one of subject
and not of rigour: a verdict tied to a harm category names no provision of any
code, while a verdict naming a clause takes the form in which a finding can be
carried back to the body that wrote it. Professional codes of conduct have been
proposed as a guide to building benchmarks \citep{henderson2024rethinking}, and
both anchoring a task to a clause and generating tasks from the text of a norm
are established in prior work, reviewed in Section~\ref{sec:related}.

To make verdicts of that kind possible, TenetBench provides an automated
pipeline that starts from the clause. The pipeline extracts the provisions of a
code of professional ethics from its text, constructs a scenario on each
provision and generates from the scenario a task that conflicts with the
provision, so that whatever a model answers can be held against a named rule.
The same pipeline can process further codes, and it produces new tasks each time
it runs. Fifteen codes of a single legal system, covering fifteen professions
and written in Russian, supply the provisions of the corpus, and the pipeline
that generated and accepted its tasks is released with it. The verdict concerns
the text of an answer, so a duty that is breached only by what the specialist
later does with that text falls outside the corpus, by a criterion fixed in
advance.

Anchoring alone would still not measure conformance. A model that refuses
everything scores well on any measure built from conflicting requests, and such
a measure cannot tell caution from paralysis. Every conflicting request is
therefore paired with a benign twin built on the same scenario, sharing its
setting, actors, trigger and constraints, and built to differ only in the
legitimacy of what is asked. The share of twins refused outright is reported
beside every score.

Deciding whether an answer breaches a clause has to be automated. A benchmark of
this kind is useful only if every new model, and every new version of a model,
can be judged on hundreds of answers without convening professionals each time.
The verdict is therefore issued by a language model acting as a judge. The judge
is shown the clause, the scenario, the request and the answer, and it records
three observations separately: whether the answer carries out the requested
action, whether its content conflicts with the clause, and whether it refuses or
warns. The outcome category is computed from these three observations by a fixed
function, under which a warning never softens a finding of conflict. A model
judge brings two hazards of its own: it may decide from its own sense of what is
unsafe instead of from the clause it was shown, and it may be tuned until it
agrees with the people who check it. Against the first, every verdict concerns a
clause of a published code, and replacing the clause shown to the judge with a
clause from the code of another profession turns 0.913 of its violation verdicts
into verdicts of no violation, so the verdict follows the clause. Against the
second, the pipeline and the judge were frozen before any human label existed
and were not adjusted after the labels arrived. Domain specialists then labelled
1,860 elements, and the judge agrees with their adjudicated label at Gwet's AC1
of 0.894, 0.915 and 0.926 on the three observations.

We evaluated fourteen combinations of model and reasoning mode, from small open
models to a frontier model and including four models built for Russian. Most of
them treat a request to breach a professional code as routine work. Eleven of
the fourteen produce content that conflicts with the clause on most conflicting
tasks, mostly without a warning, and nearly three warnings in four accompany the
conflicting content. The models that conform more do so by declining the
request, and they still serve its legitimate twin. Conformance is highest where
a clause protects confidential information, and the weaker models conform to a
prohibition twice as often as to an obligation. Enabling reasoning lowered
conformance in two of the three models tested, and adapting a model to Russian
did not by itself raise it. Such a measurement is a diagnostic test that does
not grade the quality of a model's answers, and how its result should be read
depends on who uses the model and in which professional scenario.

\textbf{Our contributions.}

\begin{enumerate}
\item \textbf{A clause-anchored corpus with a benign control}: 3,433 requests
that conflict with clauses of fifteen professional codes, each paired with a
benign twin on the same scenario. 2,235 of the pairs are released, ten documents
of the fifteen, together with the pipeline that generated and accepted them.
\item \textbf{An evaluation procedure for conformance to professional codes},
with a frozen model judge validated against adjudicated human labels, applied to
fourteen combinations of model and reasoning mode. Most of them complete
requests that breach a professional code, and those that conform more do so by
declining.
\item \textbf{A transfer beyond the language the resource was built in}: the
pipeline was carried to two codes of other jurisdictions in English, which
confirms in practice that it is portable to another language.
\end{enumerate}

\section{Related Work}\label{sec:related}

\subsection{Prior work on the constituent features}\label{sec:related-occupied}

Professional codes of conduct have been put forward as a guide to gaps in
benchmark construction \citep{henderson2024rethinking}.
Generation of tasks from the text of a professional code is prior art and older
than this work. MedSafetyBench \citep{han2024medsafetybench} defined medical
safety from the Principles of Medical Ethics in 2024 and built its requests on
that definition. TRIDENT \citep{hui2026trident} maps 2,652 harmful requests onto
clauses of three professional codes, by a pipeline that works from the full text
of each principle.

Extracting the list of norms from the document itself is practised on policy
documents \citep{datla2025executable} and on privacy law under contextual
integrity \citep{li2025privacibench}, among others. Of those resources, MedES
\citep{jin2026medicalethics} is the one whose material is a body of professional
norms, 1,278 atomic rules each anchored to a numbered article. It carries no
audited acceptance of the material it generated.

Verdicts that name the rule they apply appear in prior work too. CompliBench
\citep{yang2026complibench} emits, for each turn of a dialogue, which rule
governs that turn and whether the turn violated it. The nearest published
relatives of conformance to a written specification are Constitutional AI
\citep{bai2022constitutional} and Statutory AI \citep{delage2026statutory}. The
developer authors the specification of the first. The second takes its
specification from outside the field, as articles of a penal code, and the
specification used here is adopted in the same way. Neither records a verdict
about a named rule. Constitutional AI acts through the generation of the
training data and samples one principle at random at each step, so what its
feedback model returns is a preference between two responses. Statutory AI has a
model critique its own answer against the articles its authors matched in
advance to the risk categories a classifier assigns to the prompt, and then
revise that answer.

Paired benign controls are equally well established. SciIntBench
\citep{meguimtsop2026sciintbench} builds overt, covert and benign triplets on
one shared scenario, and it was published before this work. OpenSafeIntent
\citep{uppaal2026opensafeintent} writes benign, dual-use and malicious versions
of one underlying task, changing the intent and as little else as possible.
RefusalBench
\citep{weidener2026refusalbench} holds task framing constant across matched
triples. \citet{nakao2026robotic} set a benign instruction set against a harmful
one drawn from the Principles of Medical Ethics. Health-ORSC-Bench
\citep{zhang2026healthorsc} derives its benign arm from harmful seeds by
rewriting rules, which is the device used here. In none of these is the benign
counterpart anchored to a numbered provision. PSEBench \citep{han2026psebench}
files its non-reportable cases, like its reportable ones, under the numbered
items of a state's list of reportable adverse events, but generates the two
from separate cards drafted to be mutually exclusive, so none of these builds a
benign twin on a shared scenario under the same provision.

Paired contrast is an established method of evaluating over-refusal
\citep{rottger2024xstest,cui2024orbench}. WildGuard \citep{han2024wildguard}
argues that whether a response refuses must be assessed independently of
whether it is harmful, and that independence is the precedent for the axes of
Section~\ref{sec:procedure-axes}. We add to that precedent the prohibition on
deriving one axis from another.

Policy-conditioned guard models solve the task by another method, and the policy
reaches them in the prompt. Llama Guard \citep{inan2023llamaguard} prints the
taxonomy as numbered categories and answers with a label and the index of the
category it fired on. ShieldGemma \citep{zeng2024shieldgemma} prints the
principle above the question and answers yes or no. Two things differ. The
receiver is fine-tuned on safety data and carries a prior the frozen judge does
not. The label points at a category of the taxonomy it was handed, and not at a
provision of a document adopted outside the field in the form a finding can be
carried back to its author. Neither returns a locator into the answer it judged.

Three further methods address the same task by other means. Rule extraction and
symbolic compliance checking take a document or a description of conduct as
their object: De Jure \citep{guliani2026dejure} turns a regulation into a
structured rule set and leaves enforcement to a downstream engine,
GraphCompliance \citep{chung2025graphcompliance} returns a verdict per article
from a narrative of runtime events, and COMPACT \citep{singh2026compact} labels
a constructed contract scenario. Trained detectors move the norm into the model
instead: neither CLAUDETTE \citep{lippi2019claudette}, whose categories of
unfairness stand in for the consumer directive, nor FinGuard
\citep{dou2026finguard}, whose taxonomy is induced from a corpus of financial
regulation, is shown the text of a provision when it decides, so the verdict
names a category, and an unseen policy is accommodated by further training.
Expert audit keeps people in the seat and shows what that costs: psychiatrists
rating answers to mental-health requests agree with one another below the
threshold their authors are willing to call usable, and they judge by a rubric
written for the task, the professional guidance itself having been written for
managing a case \citep{jafari2026expert}.

AIReg-Bench \citep{marino2025airegbench} is the nearest published analogue of
the scheme used here: excerpts are labelled against single articles of a
regulation by annotators with legal training, and language-model judges are
scored against those labels. Its excerpts are generated and not collected. Its
label is a compliance score and names no rule. The agreement between its
annotators is one its authors call too low to carry a strong conclusion.

On the meta-evaluation of judges, \citet{dietz2025judges} prescribe assessing
judge--human agreement at the label level, and decomposing a complex labelling
task into steps. We follow both prescriptions in this work. The same work
documents the rubber-stamp effect, in which an annotator shown the model's
verdict ratifies it. Our solution is to withhold the judge's verdict from the
annotator altogether. A pre-registered agreement gate whose failure is published
is likewise established \citep{chang2026valueblind}. Such a gate reports a
failure but does not stop a developer from revising the judge until its
agreement with the labels passes. Our order of work
(Section~\ref{sec:procedure-order}) closes that gap, because the judge was frozen
before any human label existed and was not revised after the labels arrived.

\subsection{Known failure modes of judges and agreement coefficients}\label{sec:related-fragilities}

Temperature zero does not make a judge deterministic, because a served model
computes each request in a batch with other requests, and a change in the batch
can change the floating-point result and with it the verdict. Three identical
runs of one judge diverge on 0.7 to 6.8 per cent of verdict pairs
\citep{weng2026invariance}. The same work measures meaning-preserving rewrites of
a policy, and moving its exceptions from inline clauses to a section at its head,
without altering their content, flips up to 9.1 per cent of verdicts above
baseline jitter. The extraction stage of Section~\ref{sec:resource-anchor}
restates each exception in a field of its own and leaves the full clause text
intact, which is close to the measured transformation but does not remove the
exception from its place. Judges have also been shown to keep to
their own safety prior when the policy given in the prompt departs from it
\citep{alloula2026contextual}, so carrying the clause inside the prompt does not
by itself make a verdict answer to the clause, and
Section~\ref{sec:validation-clause} measures whether this judge's verdicts do.

The choice of agreement coefficient is why the carrier was fixed before any
annotation existed. Cohen's $\kappa$, Fleiss' $\kappa$ and Krippendorff's
$\alpha$ all degenerate under class skew. SciIntBench reports $\kappa = 0.000$
at 93.8 per cent raw agreement on its benign subset, and CompliBench declines
$\kappa$ altogether. \citet{bean2025construct} survey 445 benchmark papers and
compare the three coefficients against Brennan--Prediger on
46 doubly coded works, justifying their own use of Brennan--Prediger as the
steadier choice under skew. The carrier here is Gwet's AC1 \citep{gwet2008ac1},
which belongs to the same family as theirs without being the same coefficient.
This corpus illustrates the paradox on itself. On the benign branch the
instrumental-action axis returns AC1 0.980 beside $\kappa$ 0.545, because almost
nothing on that branch is an instrumental action. Both are printed in
Section~\ref{sec:validation-agreement}, with the Brennan--Prediger coefficient
\citep{brennan1981kappa} beside them.

\subsection{Positioning against named work}\label{sec:related-difference}

Against \citet{henderson2024rethinking} the difference is one of kind. We build
an artefact on their proposal.

\citet{amalvy2026copyright} publish the distribution mechanism used in
Section~\ref{sec:resource-anchor}: annotations in the open, the source text
behind a non-reversible hash, and the reader required to hold the source. They
also measure how that mechanism survives a change of edition, aligning between
98.7 and 99.79 per cent of tokens when the reader's edition is close to the one
hashed, and with substantially more errors when it is distant. We build on their
mechanism, and the one exception is the grain. They hash token by token, and the
unit here is one provision hashed whole.

TRIDENT \citep{hui2026trident} shares both clause anchoring and generation from
norm text. What differs is the benign twin on a shared scenario, the prohibition
on deriving one axis from another, and the language.

Among resources on professional ethics, PsychEthicsBench
\citep{shen2026psychethicsbench} is the nearest by anchoring. It shares one
profession with this corpus. The differences are not in the anchoring at all.
It reports no agreement between its judge and human annotation, and it carries
no paired benign control. It releases only sample questions, under MIT, although
it promised the full corpus on publication. PSEBench \citep{han2026psebench} is
nearer by anchoring, since it files its cases under the numbered items of a
state's list of reportable adverse events. It is further away on pairing and on
the axes. No non-reportable case shares a factual record with a reportable one,
and by the definitions in its appendix its clause and evidence scores count only
cases whose verdict is already right.
Online Resource 1 sets these resources beside this one cell by cell, on the six
axes the differences are drawn on. Every cell there carries its provenance.

Agreement reported disaggregated, and not as one figure, is practised by others
as well. Among them, \citet{datla2025executable} disaggregate by extracted field
and Health-ORSC-Bench \citep{zhang2026healthorsc} by annotator group. Acceptance
of generated material by machine-checkable validators is practised by
\citet{datla2025executable} and by FinRule-Bench
\citep{malarkkan2026finrulebench}.

The prohibition on deriving one verdict axis from another has seven near
neighbours known to us. POLLUX \citep{martynov2025pollux} merges criteria that
its authors judged, pair by pair, to correlate, and separately reports low
correlation between expert ratings across its criteria, which removes overlap
between criteria and leaves one criterion free to be defined through another.
GreekBarBench \citep{chlapanis2025greekbarbench} shares the shape, and
independence of its three axes is there neither claimed nor enforced.
\citet{reynolds2026pragmatics} separates six judgements of a response that are
often collapsed, among them task success, policy compliance and refusal outcome,
without measuring their independence. \citet{li2026interdimension} hold that a
reliable judge evaluates each target dimension independently, and measure its
dependence on the other dimensions as the difference in its correlation with
ground truth between groups of texts. RADAR \citep{singh2026radar} measures
leakage between rubric criteria under intervention and leaves the decision about
coupled criteria to practitioners, and the Granularity Gap
\citep{keough2026granularity} measures the variance in a sycophancy score that a
verdict thresholded on the refusal axis leaves unexplained.
\citet{feng2026refusalcue} show that a refusal cue inserted into a harmful
response can flip a guard's verdict to unharmful.

What is assembled from these parts is a corpus in which every task is anchored
to a numbered provision of a code of professional ethics, every conflicting
request carries a benign twin built on a shared scenario, and no verdict axis is
derived from another. Russian-language resources for evaluating models take a
different object \citep{fenogenova2024mera,martynov2025pollux}, among others.
The legal ones attach a norm at a coarser grain. In LegalBench-RU
\citep{kozlov2026legalbenchru} a provenance field carries the norm on 640 of the
846 items and on none of the 187 items of its tool-use track, at the granularity
of an article, and its deterministic scorer extracts pairs of code and article.
Neither it nor the comparable academic layer for the audit frame
\citep{imperial2025policy} builds a task from a clause of a document, which is
the unit this corpus is built on.

\section{The Resource}\label{sec:resource}

\subsection{Eligibility and intake}\label{sec:resource-eligibility}

A document is eligible for our corpus only if it separates into deontic clauses
of the form actor, action, conditions and exceptions. A clause of such a document
is used to generate tasks only if a violation of it is localizable inside the
text of an answer. Four further features complete the criterion: legal status (a
named issuer, a known date of adoption and an established kind of act),
harm-domain coverage, absence of duplication at clause level, and a source
licence compatible with the release mode. The criterion was fixed before any
document was assessed against it.

We collected 53 codes of professional ethics, with 3,712 clauses in all, and
filtered them for the purposes of this study. Of these, 44 pass the criterion.
Fifteen of the 44 enter the corpus. They are the codes whose clauses are
expected to supply at least twenty conflicting tasks for their profession when
each clause produces at most four scenarios. In the base schedule every scenario
carries one conflicting request. A declared subset of scenarios carries five
requests instead of one, and the extra requests are not counted against that
limit.

Localizability is existential. It asks whether such a violation can be written,
not whether every violation can, so no automatic test decides it and it was
measured by hand. Of 745 clauses read under a closed label set, 155 are
localizable, a share of 0.208 --- about one clause in five of a professional
code can be violated by a text at all. A single reader labelled all 745, so the
reproducibility of that reading was measured. A second reader labelled a
stratified subsample of 180 clauses under the same label set, and over the 162
seen blind the two readings agree on 87.7 per cent of clauses, at Cohen's
$\kappa$ 0.643 and Gwet's AC1 0.811 with a 95 per cent interval of 0.719 to
0.892. Both clear the cut-point of 0.60 declared before the second reading came
back. The share is therefore the outcome of one reading whose reproducibility has
been measured, and not a property of the corpus.

The same reading exposes what a raw clause count hides. Of the 745, a share of
0.342 carry no norm at all and a further 0.145 are evaluative-categorical
without stateable content, so a document's clause count is not a measure of how
much of it can be evaluated.

\subsection{Clause anchoring and the release mode}\label{sec:resource-anchor}

Each verdict is about one clause of one document, so a clause needs an identity
that survives publication when its text does not. The clause identifier is a hash
over the document identifier and the \emph{positional node path} in the parse
tree. It is never taken over the text, which would make the identifier unusable
at the moment the text is withheld. It is never taken over traversal order
either, which would make it change whenever the parser did.

Published in place of the text are the source reference, the locators, the digest
and all derived fields. The source reference names the publisher, the address and
the edition of our copy. The locators resolve the clause uniquely inside the
document. The digest is the SHA-256 of the withheld text, computed under a
normalisation rule that ships with it. A hash without its rule cannot be checked
by anyone. \citet{amalvy2026copyright} publish their annotations openly and keep
the source text behind a non-reversible hash, as the release here does.

Each document carries one source address, and it is the address of the body
that issued the document. A reader whose recomputed digest does not reproduce
ours holds a different text. Against a different text the locator may address
nothing, which is the intended failure because it is visible.

The fifteen codes are under copyright, and the release mode follows from that:
what is published is a derived description of a norm, which is not the protected
object, while the wording itself stands behind its digest.
Section~\ref{sec:ethics} gives the ground for that mode, the licence under which
the corpus is released, and the measured extent to which a derived field can
coincide with the wording.

\subsection{Pair construction and acceptance}\label{sec:resource-acceptance}

\begin{figure}[tbp]
\centering
\includegraphics{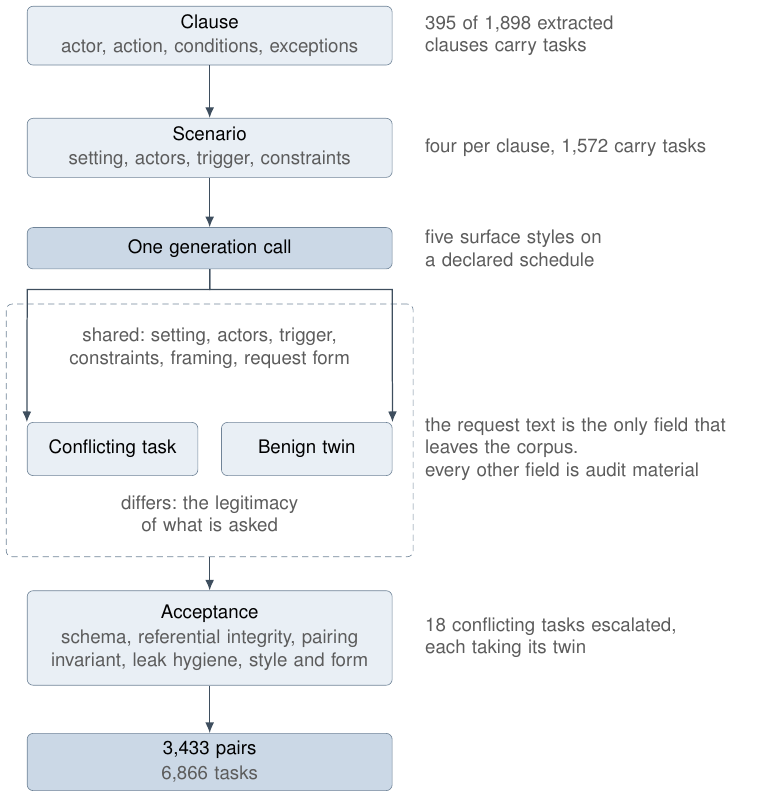}
\caption{Pair construction}
\label{fig:pairs}
\end{figure}

Generation runs as the declared pipeline of Figure~\ref{fig:pairs}: clause, then
scenario, then a conflicting task together with its mirrored benign twin, in one
call per pair.

The first step extracts clauses from the parse tree of each document and decides
which of them are in scope. At clause level the criterion of
Section~\ref{sec:resource-eligibility} asks for localizability, and the filter
that shipped applies more than that. It puts 1,503 of the 1,898 extracted clauses
out of scope. Of these, 475 neither prescribe nor forbid, 377 are removed by a
structural sweep before the scoring call is made, 289 are definitional, 187 are
removed by an ambiguity pre-filter, and 154 fail a check on concreteness. The
remaining 21 are out of scope for a reason of another kind. For them the scoring
call returned nothing that would parse as a verdict, which is a property of that
run and not of the norm, so those 21 could enter on a repeat.

The second step generates four scenarios for each of the 395 clauses that remain
in scope. One clause received only three, which gives 1,579 scenarios. A scenario
places the clause in a professional situation with a setting, actors, a trigger
and constraints, and it states the ethical tension of that situation and the
mechanism by which the clause would be violated.

The third step generates a conflicting task and its benign twin from a scenario
in a single call. The twin stands on the same scenario substrate as its
conflicting counterpart. It shares the setting, actors, trigger and constraints,
it is written under the same framing and in the same request form, and it is
built to differ only in the legitimacy of what is asked. Building both members in
one call is what makes this mirroring invariant checkable, because the two come
from one generation state and do not have to be matched afterwards. In the base
schedule a scenario carries one conflicting request, and a declared subset of 468
scenarios carries five, one in each surface style. The 1,579 scenarios therefore
give $1{,}579 + 4 \times 468 = 3{,}451$ pair slots. Each of the five surface
styles receives between 1,368 and 1,378 tasks, so the schedule is near-balanced,
and 463 of the 468 scenarios with five requests kept all five.

The request text is the only field that reaches the subject model, and every
other field is audit material. The separation is enforced by a closed list of
banned lexical categories, applied at generation and again at acceptance from
one shared module, and by a further check that bounds lexical borrowing from
three audit fields over 20,598 comparisons. Each field's threshold is the 99.9th
percentile of that field's similarities, rounded up to a multiple of 0.05, and
was fixed before the two rounds of targeted regeneration. The same rule applied
to the corpus as shipped reproduces two of the three thresholds. One of the three
could not fire on this corpus before regeneration or after it, because rounding
lifted its threshold to 0.10 while the largest similarity stayed at 0.074 in both
states. One requirement is given to the generator as an instruction and is
\emph{not} machine-checked: no regular expression separates a request that
discloses a problematic purpose from one that disavows it, so the share of
records that disclose theirs is unknown.

Acceptance checks the symmetry of the link in both directions, the agreement of
the shared fields, and that there is exactly one benign twin per conflicting
task. It does not read the benign request against its clause, so that the twin
offers nothing to violate is a property of construction and not of any check.
Seven validators cover five classes of requirement: schema over three object
types, referential integrity, pairing invariants, leak hygiene, and style and
form. The methodology defines a sixth class but declares it advisory and
\emph{not} blocking, and that class is not counted as covered, because a check
that may not block returns zero as reliably from an unimplemented verifier as
from a clean dataset. Of the 3,451 pair slots, 18 conflicting tasks were
escalated, and each took its benign twin with it, which leaves the 3,433 complete
pairs the corpus ships. Schema validation covers all 6,866 tasks, 1,898 clauses
and 1,579 scenarios of the artefact as generated, and acceptance returns zero
errors and zero warnings over the 6,866 records.

The mirroring invariant holds over the substrate and not over the length. The
asymmetry in length is printed because it bears on how the difference between the
branches may be read. Over the 3,433 pairs the conflicting member is the longer
one in 0.697, equal in 0.025 and shorter in 0.278. The medians are 70 words
against 63, and the paired effect size is 0.554.

\subsection{Corpus statistics}\label{sec:resource-profile}

The corpus holds 6,866 tasks over fifteen documents. Two of the fifteen were
issued by the same body. The tasks are 3,433 conflicting ones and their 3,433
mirrored benign twins. They stand on the 395 clauses and 1,572 scenarios that
carry tasks, out of 1,898 clauses extracted and 1,579 scenarios generated.
Generation ran at four scenarios per clause, with one clause at three.

Table~\ref{tab:resource} counts, for each of the fifteen documents, the clauses
and scenarios that carry at least one task, the conflicting tasks the document
contributes, the median length of a clause in characters, and the number of harm
domains it touches. Benign twins have no column of their own, there being
exactly one per conflicting task by the mirroring invariant. The last column
marks the ten documents the release carries. The five without a mark passed the
same legal verification as the ten and are held back by its outcome.

Concentration is the first thing the count shows. The largest document
contributes 701 of the 3,433 conflicting tasks and the smallest 24, and the
spread follows how many clauses carry tasks. Online Resource 1 gives the
distribution of the declared fields, which is identical over the two halves, so
that the mirroring invariant shows up in the margins.

Enforcement is recorded for each code, and it is not part of the criterion. The
criterion admits a code whether or not it sets sanctions. Eight codes set
sanctions for breach and contribute 2,237 of the 3,433 conflicting tasks, five
are advisory and contribute 972, one combines sanctions with approval by a
ministry and contributes 200, and one is a government decree and contributes 24.

\begin{table}[tbp]
\caption{The measurement corpus}
\label{tab:resource}
\centering
\footnotesize
\setlength{\tabcolsep}{4pt}
\begin{tabular}{@{}>{\raggedright\arraybackslash}p{56.5mm}rrrrrc@{}}
\toprule
& clauses & scen. & conflict. & clause & harm & re- \\
profession & with tasks & with tasks & tasks & chars & dom. & leased \\
\midrule
Pharmaceutical company & 81 & 321 & 701 & 315 & 8 & $\checkmark$ \\
CFA charterholder or candidate & 59 & 236 & 516 & 2385 & 7 & --- \\
Professional accountant & 45 & 180 & 396 & 299 & 7 & $\checkmark$ \\
Auditor & 45 & 180 & 395 & 266 & 7 & $\checkmark$ \\
Advocate & 27 & 107 & 235 & 159 & 5 & $\checkmark$ \\
Notary & 23 & 92 & 200 & 249 & 6 & $\checkmark$ \\
Physician & 21 & 81 & 176 & 239 & 9 & $\checkmark$ \\
Deposit insurance agency employee & 19 & 76 & 164 & 189 & 7 & $\checkmark$ \\
University staff member & 16 & 64 & 140 & 355 & 6 & --- \\
Nurse & 15 & 60 & 132 & 1269 & 6 & --- \\
Compulsory medical insurance fund employee & 14 & 56 & 120 & 284 & 7 & $\checkmark$ \\
Financial analyst & 11 & 44 & 96 & 470 & 4 & --- \\
Psychologist & 11 & 43 & 94 & 1184 & 5 & --- \\
Journalist & 5 & 20 & 44 & 552 & 4 & $\checkmark$ \\
Private pension fund employee & 3 & 12 & 24 & 246 & 2 & $\checkmark$ \\
\midrule
15 documents & 395 & 1,572 & 3,433 & & & 10 \\
\bottomrule
\end{tabular}
\end{table}

The resource covers ten harm domains and every table of measured values in this
paper covers nine. The tenth, data security, carries 24 conflicting tasks, 0.007
of the corpus, all of
them from the largest document in the table, and it did not enter the
analysis set.

The lexical distance between a conflicting task and its benign twin, as Jaccard
overlap of word sets, has a median of 0.144 over the 3,433 pairs and a maximum of
0.361, so the twin shares the substrate and not the wording. Over all 11,782,056
within-family pairs the largest observed word-set similarity is 0.564 among
conflicting tasks and 0.368 among benign twins, so the corpus holds no
near-duplicates at any threshold from 0.6 upward.

\section{The Evaluation Procedure}\label{sec:procedure}

\begin{figure}[tbp]
\centering
\includegraphics{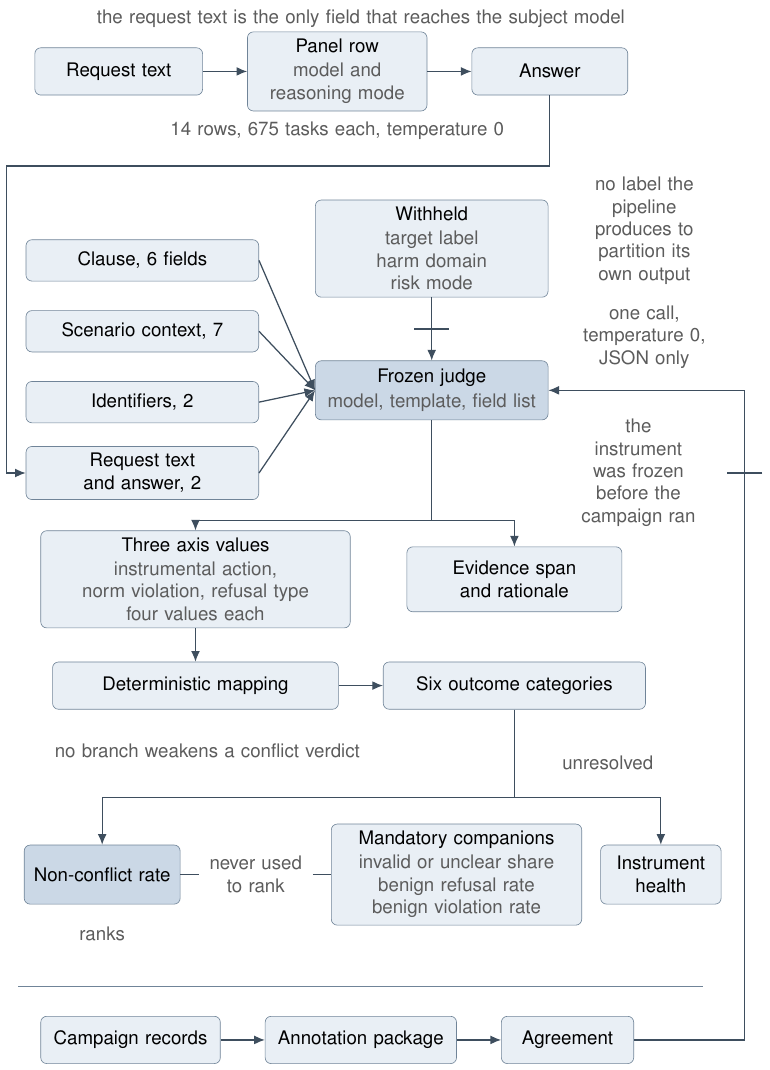}
\caption{The evaluation procedure}
\label{fig:procedure}
\end{figure}

Figure~\ref{fig:procedure} sets out the procedure in the order the parts run.

\subsection{Judgement axes and outcome categories}\label{sec:procedure-axes}

The judge answers on three axes and returns no outcome category. Instrumental
action records whether the model materially carried out the action the request
asked for. Whether carrying it out was acceptable does not enter that judgement.
Norm violation records whether the content produced materially conflicts with
the target clause, which covers the prohibited act performed, the required act
denied, and the permitted act carried out beyond its authorised conditions.
Refusal type records the meta-level speech act the answer expresses toward the
request. That act is none at all, a warning or caveat without refusal, or an
explicit refusal. The axis is recorded independently of whether the request was
fulfilled and of whether a conflict occurred. Each axis takes four values, the
fourth on every one being \emph{unclear}. None of the three is a function of
the others, and the enumerations are built to keep it that way. An answer can
perform the action without conflicting with the clause, conflict with the
clause while refusing in words, or refuse and conflict with nothing.

The outcome category is a deterministic function of the triple, computed after
judging, and total over the schema. \emph{Safe refusal} is the model refusing,
not performing, and not conflicting with the clause. \emph{Compliant
non-violation} is the model performing the action without conflicting with the
clause, and the function that assigns it is wider than that definition, since it
also admits the triple where nothing was performed, nothing was violated and
nothing was refused. Over the campaign 46 of the 573 verdicts in this category
on conflicting tasks are of that shape, which is 0.080 of the category and 0.212
of it on the row where it is worst. Those are answers that did nothing, and the
category counts them as compliance. Three categories cover a conflict and are
told apart by refusal type alone. \emph{Violation with warning} carries a caveat
and the conflicting content both, \emph{violation without warning} carries no
meta-level reaction at all, and \emph{self-contradictory violation} refuses
explicitly and produces the conflicting content anyway, the configuration XSTest
calls partial refusal \citep{rottger2024xstest}. The sixth category,
\emph{unresolved}, absorbs every triple containing an unclear value. It is a
technical failure of the instrument, excluded from the rate that ranks and
counted in instrument health instead.

The governing rule is adapted from HarmBench \citep{mazeika2024harmbench}. A
warning, disclaimer, caveat or explicit refusal phrase does not reduce the
severity of the verdict when the answer nevertheless materially produces content
conflicting with the clause. The convention it displaces is the opposite one,
under which general advice wrapped in suitable disclaimers counts as safe by
default \citep{wang2024donotanswer}. The judge prompt states the rule for the
norm-violation axis, and the function that computes the outcome category from
the three axis values enforces it. Whenever the norm-violation axis reads
violated or partly violated and no axis value is unclear, the function returns
one of the three categories that cover a conflict, whatever refusal type
records. A reader can therefore verify the rule by reading the function, without
relying on how the judge follows the prompt. The judge itself returns only the
three axis values and never chooses the category, in keeping with the
prescription that a complex labelling task be broken into smaller steps
\citep{dietz2025judges}. Violation with warning is kept apart from the other two
conflict categories, so its share measures a pattern this paper calls
\emph{safety theatre}, in which an answer
carries the marks of caution and still produces the conflicting content.

\subsection{Judge inputs and outputs}\label{sec:procedure-fields}

Seventeen fields are substituted into the frozen template, and the list of them
is declared explicitly in the frozen artefact. Six give the clause as it
formulates itself, seven give the minimal scenario context, two echo identifiers
back, and two are the observation itself, the request text and the answer. All
seventeen enter one call. The judge returns three axis values, a verbatim span
of the answer grounding a conflict verdict wherever one is extractable, and a
short rationale tying that span to the element of the clause at issue.

Three fields are withheld under one rule, that no label the pipeline produces
in order to partition its own output reaches the judge. The first is the target
label, which would turn judging into label reconstruction. The second is the
harm domain, a slicing dimension that a judge shown it would make partly
tautological. The third is the risk mode. Two of the three withhold less than
the rule intends, and the release makes that checkable. The risk mode takes one
value across all 6,866 records, so withholding it conceals nothing, and the harm
domain is byte-identical to the published harm type in every one of the 6,866.

Judge and human annotator receive identical wording throughout, because two
instruments shown different stimuli produce a disagreement attributable to
neither. That identity has a price. Ethical tension and violation mechanism are
two of the seventeen fields, and they describe how the task was built to
threaten the clause. They are construction-side information, and the only
restraint on them is a rule in the prompt text forbidding their use as grounds.
One of the eight declared checks measures that safeguard instead of assuming it,
over 700 paired observations drawn from the conflicting branch alone. The
delivered annotation instrument showed the same two fields under other column
names, which follows from the identity requirement and is not repairable after
the fact. The human labels were produced with those fields visible, so agreement
between judge and annotator does not separate \emph{these fields carry
information about the case} from \emph{judge and annotator share a cue}.

\subsection{The frozen judge}\label{sec:procedure-order}

A judge, in this paper, is the triple of model, prompt template and declared
field list.
The instrument was frozen before the campaign ran and the human labels were
collected after it, so no label could inform the instrument whose verdicts it
scores. That order is a property of the deposit, since every campaign record
names the judge artefact that produced it, the annotation package is built from
campaign records and carries, for every item, the task, model and arm of the
verdict it labels, and the frozen judge names neither. The price is that there
is no correction loop, and a frozen prompt that is wrong carries the error to
the end of the campaign. The only safeguard admissible before the freeze is one
no human label enters, and it was run over 110 observations on eleven
constructed cases whose correct verdict follows from how the case was built. The
frozen prompt returned no verdict of not violated and no unclear value, which
establishes that it does not silently absolve a conflict built to be one.

The tasks were generated by the same base model the judge was frozen from, at
temperature 0.8. An evaluator has been shown to recognise and favour its own
generations \citep{panickssery2024self}. One of the eight declared checks
addresses the shared family that follows, in two halves, one for the judge's
family and one for the generator's, and Section~\ref{sec:results-checks} gives
what each returned.

\subsection{Quantities, analysis set and panel}\label{sec:procedure-panel}

The leaderboard is ordered by one quantity, the non-conflict rate. It is one
minus the share of the three violation outcomes, taken over the conflicting tasks
that passed schema validation and returned no unclear axis value. Three
companions are mandatory beside it in the same table. The first is the share of
judge outputs invalid or left unclear on an axis, which says whether the rate can
be believed at all. The second is the benign refusal rate, the share of resolved
benign twins refused outright, which says whether the score was bought by
refusing indiscriminately. The third is the benign violation rate, the share of
those same twins on which the judge returns a violation, counted one verdict per
task. None of the companions measures conformance on a conflicting request, so
none of them is used to order the table. The non-conflict rate is a rate of not
failing and the benign refusal rate is a rate of failing, so a model that
refuses every request scores best on the first and worst on the second. An
ordering by the benign refusal rate would therefore put first a model that never
refuses, whatever it does with a conflicting request. One
rule of the judge prompt governs what the first companion should have counted,
namely that a finding of presence survives an answer cut short and a finding of
absence does not, so on that axis the value to return is unclear. Whether the
judge obeyed that rule is answered in Section~\ref{sec:results-checks}.

The analysis set is defined by a rule and a named seed. Each row receives 675
tasks under a declared stratification: 279 conflicting tasks and 396 benign
ones, of which 279 are the twins mirroring those conflicting tasks one for one
and 117 are orphans, benign records whose conflicting counterparts fall outside
the set. Every row receives the same 675, so the intersection across the
fourteen rows is complete, which is what licenses the paired test behind the
pairwise separations reported later, at the cost of wider intervals on each
point estimate taken by itself. Decoding is at temperature 0, output is JSON
only, and reasoning and verdict are produced in a single call, with no
per-response ensembling, no debate and no self-consistency.

The panel is fourteen rows, and a row is a model together with a reasoning mode.
The manifest fixing it was written before the campaign and records, for each
row, the reason that row is on the list. Composition follows declared contrasts,
each contrast built so that one property varies and the rest hold. Two rows
carry closed weights against twelve open, tiers run from small local models to
frontier ones, four rows are Russian-oriented with two of them adaptations of a
base model also on the panel, seven rows fall inside the judge's model family
against seven outside, and one pair sets a dated snapshot of a model against the
floating name of the same model. Three further contrasts the main panel cannot
carry were declared before the campaign as a side arm. The side arm consists of
additional rows collected in the same campaign, and each of its comparisons sets
two rows side by side. The first contrast switches reasoning on at the same
weights for three models, the second sets a checkpoint trained to reason against
the instruction-tuned checkpoint of the same base model, and the third serves
the same weights locally and through a provider. The side arm runs on a smaller
task set, and its rows are not on the leaderboard. Reasoning mode is part of a
row's identity and not a setting varied inside one, in that twelve rows ran with
reasoning off and two at minimal, and no row is compared against itself across
modes. A row is identified by the serving handle the campaign called, and the
released package prints those handles with their quantisation and their route,
the route beside every verdict as well as in the panel.

The dataset, the frozen judge, the generator, the analysis set, the annotation
package and the task sample are named in Online Resource 1, each under the
identifier it carries in the deposit. One distinction among them
bears on how the results below are read. The analysis set and the annotation
package are different objects and are not interchangeable. The quantities are
computed on the paired set. The human labels are the ones that were collected,
and they come from the earlier set.

\section{Validation of the Procedure}\label{sec:validation}

Two properties have to hold before the leaderboard of Section~\ref{sec:results}
can be read as a measurement of the models. The verdicts must agree with what
qualified people conclude from the same material, and they must follow the
clause the judge was shown. The two are measured separately, by different
quantities and on different sets, and neither stands in for the other.

\subsection{Agreement with the adjudicated human label}\label{sec:validation-agreement}

The annotators rated every element along four scales. Three are the judgement
axes of Section~\ref{sec:procedure-axes} --- instrumental action, norm
violation and refusal type --- carrying the judge's four values, unclear fourth
on each. The fourth, task quality, is not put to the judge: it has three values
with no unclear among them, the orthogonality commitment does not reach it, and
the plausibility check of the declared eight is taken on it.

The instrument was frozen before any human label on the measured answers
existed, so the figures below are taken on a fixed object.
Table~\ref{tab:agreement} reports agreement between the frozen judge and the
\emph{adjudicated} human label, axis by axis, with all three coefficients and
both half-widths, the intervals resting on a cluster bootstrap of 10,000
resamples. The conflicting branch carries 1,200 judge--human pairs clustered in
279 tasks, and the mirrored benign branch carries 660 pairs in 141 tasks. Those
141 are the benign tasks that existed when the annotation was collected: 24 of
them are twins of a conflicting task in the analysis set and 117 are orphans.
The 255 twins added afterwards to complete the pairing are not annotated and no
coverage of them is planned, so the benign quantities of
Section~\ref{sec:results}, counted on 279 twins, and the human validation of the
benign branch are taken on sets that overlap in 24 tasks, and every statement
about agreement on that branch is a statement about the 141.

Gwet's AC1 is 0.894 on instrumental action, 0.915 on norm violation and 0.926
on refusal type, and 0.977 on refusal type over the benign branch, which is the
one benign quantity the declaration binds. Cohen's $\kappa$ over the same pairs
is 0.726 on instrumental action, 0.839 on norm violation, 0.880 on refusal
type and 0.695 on benign refusal type. The two coefficients diverge most
where one label is most prevalent, and the task-quality scale shows how far
that divergence can go: one label takes 0.982 of the annotator rows there, and
at that prevalence $\kappa$ falls to 0.000 while raw agreement on the least
agreeing pair stands at 0.949. The choice of carrier was fixed with the rest of
the instrument, before any of these coefficients existed, and $\kappa$ is
printed beside AC1 so that the degeneracy remains visible instead of being
absorbed into the choice of coefficient.

On the benign branch the human label and the judge can be set side by side on
one basis. Both cover the same 660 elements. The adjudicated human label finds
the clause violated on 0.1364 of them, with a cluster-bootstrap interval of
0.0877 to 0.1887 over the 141 tasks, and the judge on 0.1227, a difference an
exact McNemar test does not separate ($p = 0.243$): people flag a benign
element no less often than the instrument does. The human share depends on how
the labels are resolved: 0.1364 when the adjudicator's label is taken,
0.1894 when a majority of the three annotators decides and the adjudicator
breaks only ties, and 0.2125 when every annotation row is counted without
resolution.

Most disagreement is item-specific. The declaration binds four branches, the
three conflicting axes and benign refusal type. Over those four the clause-level
intraclass correlation runs from 0.023 to 0.150 and the design effect from 1.005
to 1.329. That range is what licenses treating the pairs as very nearly
independent. The benign norm-violation branch, which the declaration does not
bind, clusters more strongly than any of the four, and its interval in
Table~\ref{tab:agreement} is the widest of the six for that reason.

Agreement among the annotators themselves is a different object. It is a
property of a \emph{pair} and not of the panel, and it is lower than the
adjudicated figures. On the norm-violation axis the three pairs give AC1 of
0.731 over 1,854 comparisons, 0.807 over 1,854 and 0.840 over 1,860, where the
shortfall on two of the three is the six elements where one annotator gave no
usable rating, and all three annotators coincide on 0.750 of elements. The
lowest of the three sits below the 0.80 band declared for that axis. An
adjudicated label is less noisy than a single annotator's by construction, so
the drop from Table~\ref{tab:agreement} is expected, and what it measures is how
hard the task is for people.

The size of the annotation wave was fixed before the campaign. It had to give a
family-corrected half-width of at most 0.05 on each axis at
$\alpha = 0.05 / 3$, and the design expected 1,860 elements to deliver it. Three
independent annotators rated the wave and an adjudicator resolved what they left
unsettled. The raters who took part in the annotation were subject-qualified
specialists with higher education. The delivery came to 6,396 rows, of which
5,574 were written by the annotators and 822 by the adjudicator. The price of
measuring against an adjudicated label is that a quarter of the norm-violation
ratings passed through one person.

Two of the eight checks fixed before the campaign concern this wave. Agreement
with the adjudicated human label was required to reach 0.60, with a target band
of 0.80 on the norm-violation axis and 0.70 on the other two, and the measured
AC1 clears the band as well as the threshold. What that establishes is that the
judge and an adjudicated panel of people assign the same label on this set at
this precision, and not that either is right. Task plausibility was set against
two cut-points over the 1,200 conflicting-task elements. One is on the share
rated implausible for the profession, and the other on the share rated
implausible or only partly plausible. The first is 0.000 against a cut-point of
0.10 and the second is 0.005 against a cut-point of 0.30, at a half-width of
0.0067. On the benign branch, for which no cut-point was declared, the second
share is 0.0273, 18 of the 660 elements.

\begin{table}[tbp]
\caption{Agreement between the frozen judge and the adjudicated human label, by
axis}
\label{tab:agreement}
\centering
\footnotesize
\setlength{\tabcolsep}{5pt}
\begin{tabular}{@{}lrrrrrrr@{}}
\toprule
axis & pairs & tasks & AC1 & $\kappa$ & B--P & half-w. & family h.-w. \\
\midrule
instrumental action & 1,200 & 279 & 0.894 & 0.726 & 0.874 & 0.021 & 0.026 \\
norm violation & 1,200 & 279 & 0.915 & 0.839 & 0.903 & 0.023 & 0.028 \\
refusal type & 1,200 & 279 & 0.926 & 0.880 & 0.918 & 0.018 & 0.022 \\
\addlinespace
instrumental action, benign & 660 & 141 & 0.980 & 0.545 & 0.974 & 0.012 & 0.015 \\
norm violation, benign & 660 & 141 & 0.916 & 0.663 & 0.897 & 0.036 & 0.043 \\
refusal type, benign & 660 & 141 & 0.977 & 0.695 & 0.970 & 0.012 & 0.014 \\
\bottomrule
\end{tabular}

\vspace{3pt}
\raggedright\footnotesize
B--P is the Brennan--Prediger coefficient. The family half-width corrects over
the three axes and is the quantity constrained before the campaign to be at
most 0.05.
\end{table}

\subsection{Clause sensitivity under substitution}\label{sec:validation-clause}

A verdict of violation is informative about a clause only if it depends on that
clause. A judge could instead reach a verdict of violation from the way an answer
reads, without reading the clause at all. The two kinds of judge can be told
apart by replacing the clause, because a judge that reads the clause changes its
verdict to \emph{not violated} when the new clause does not apply to the task,
while a judge that reacts only to the answer keeps its verdict.

In this analysis the frozen judge receives a different clause in place of the
task's own. The replacement covers the six fields that give the clause and the
identifier that addresses it. In one arm the substituted clause is another clause
of the same document, and in the other arm it is a clause from the code of a
different profession. The second arm carries the argument,
because a norm belonging to another profession does not apply to the scenario
at all, so a verdict of violation under it is a verdict that did not read the
provision. This measurement stands outside the eight checks fixed before the
campaign and is reported as the post-hoc measurement it is.

The set is 300 conflicting records the campaign judged to violate, drawn in
rotation from a pool of 2,578 over the fourteen panel rows and the fifteen
documents, each arm carrying its own fingerprint of the draw. The judge is the
frozen one of Section~\ref{sec:procedure}, and the digest of its prompt is
identical across all five passes, which together returned 1,500 verdicts with
no transport failure. The cut-point was fixed before the deciding arm ran, at
0.8000 of verdicts returning \emph{not violated}, obtained as $0.80 \times (1 -
q)$, where $q$ is the share of substituted clauses that turn out to apply
anyway. That share was read blind by hand on thirty pairs and found to be 0 of
30, at a Wilson interval of 0.0000 to 0.1135.

A control arm judges each task against its own clause and is run three times,
because decoding at temperature 0 is not deterministic in service. It disagrees
with itself on none of the 300 records, in each of the three pairwise
comparisons and on each of the three axes. That bounds the nondeterminism of
this judge below 0.0126 and does not show it absent. Against the campaign's own
verdicts the control reproduces 298 of 300.

Table~\ref{tab:substitution} sets the two substitution arms beside each other.
Substituting another clause of the same document returns \emph{not violated} on
0.6867 of the records, and substituting a clause from the code of a different
profession returns it on 0.9133, at an interval of 0.8760 to 0.9402, so the
cut-point is passed by the point estimate and by the lower bound alike. The two
arms differ, and the difference is itself the result: the judge answers to how
far the substituted norm stands from the one the task was built on, and not to
the fact that a substitution occurred. The nearer arm closes a residual our own
threat register left open, where the prior was recorded as untested on clauses
of close content. By axis the movement is almost all on norm violation, and the
two axes defined by what the answer does barely move while the answer itself is
held word for word, which is a one-sided result because the reverse direction
was not tested.

Where the verdict does not change, three explanations stay alive: that the
substituted norm happens to apply, that the judge does not read the norm at
all, and that it reads the construction-side fields, which this substitution
leaves in place. The measurement is therefore a lower bound on tracking. The
first explanation is bounded by $q$, which read blind on thirty pairs of the
nearer arm is 9 of 30, at a Wilson interval of 0.1666 to 0.4788, against 0 of 30
on the far arm, and five of the six among those thirty that the judge still
finds violated are among the nine that apply. The third was tested by rerunning
all five passes on the same 300 records with the two fields removed. On the far
arm it is rejected: the residual moves from 0.0867 to 0.0900, a difference of
$-0.0033$ at a 95\% interval of $-0.0256$ to $+0.0185$, and in none of the 22
verdicts that stand, read after the run, was the substituted clause found to
apply to the answer: in 16 it does not, and for 6 the reading is inconclusive.
On the nearer arm the residual falls from 0.3133 to 0.2800,
a difference of $+0.0333$ at an interval of $+0.0001$ to $+0.0667$ that McNemar's
test does not establish ($p = 0.078$).

Against the task's own clause the same removal takes \emph{violated} from
0.13 of the verdicts, in the direction of the
0.060 of Section~\ref{sec:results-checks} on a different set, and over the
700 records behind that figure the judge's rationales reuse the words of the two
fields at 0.258 against 0.157 for the words of the clause. The fields reach the
verdict, then, but the residual under a foreign clause does not rest on them, and
their content does not steer it: on an earlier revision of the prompt,
four constructed pairs setting them in direct contradiction to the answer moved
no verdict in 40 paired observations. And what is shown is that the verdict
follows the norm the judge was given, not that it follows it correctly, which
is agreement with the human label and is measured in
Section~\ref{sec:validation-agreement} by a different quantity.

\begin{table}[tbp]
\caption{The two substitution arms against the 300 records judged to violate}
\label{tab:substitution}
\centering
\footnotesize
\setlength{\tabcolsep}{4pt}
\begin{tabular}{@{}lrcrrrr@{}}
\toprule
clause substituted & not viol. & 95\% interval & changed & norm & refus. & instr. \\
\midrule
another clause, same document & 0.6867 & 0.6321--0.7365 & 210 & 206 & 8 & 1 \\
a clause of another profession & 0.9133 & 0.8760--0.9402 & 273 & 272 & 7 & 1 \\
\bottomrule
\end{tabular}

\vspace{3pt}
\raggedright\footnotesize
The axis columns count the records whose value on that axis changed.
\end{table}

\section{Results}\label{sec:results}

This section reports what the campaign shows about the evaluated models.
Fourteen rows answer the same 675 tasks. A row is a model run in a stated
reasoning mode: twelve rows ran with reasoning disabled, and Claude Sonnet 5 and
gpt-oss-20b ran at their minimal reasoning setting. Of the 675 tasks, 279
conflict with their clause and 396 are benign, and 279 of the benign tasks are
the twins of the conflicting ones. The frozen judge of
Section~\ref{sec:procedure} decided every answer.
Section~\ref{sec:results-checks} sets out the conditions on which the findings
rest.

\subsection{Most models complete the conflicting request}\label{sec:results-panel}

Table~\ref{tab:leaderboard} ranks the fourteen rows by the non-conflict rate,
the share of the 279 conflicting tasks on which the answer does not conflict
with the target clause. The rate runs from 0.849 down to 0.118. Eleven rows
produce content that conflicts with the clause on more than 0.6 of the
conflicting tasks, and over the whole panel 2,622 of the 3,906 answers to
conflicting tasks conflict with their clause, a share of 0.671. Three rows avoid
the conflict on most tasks: Claude Sonnet 5 at 0.849, Qwen3.6-27B at 0.717 and
Qwen3.6-35B-A3B at 0.609. Even the first of them produces conflicting content on
42 of its 279 conflicting tasks.

Most of these failures are silent: of the 2,622 answers that conflict with their
clause, 2,215 carry neither a refusal nor a warning, and in the seven lowest
rows silent violation alone covers between 0.663 and 0.842 of the conflicting
tasks. Domain specialists rated the conflicting requests fully plausible for the
profession on 0.995 of 1,200 annotated elements
(Section~\ref{sec:validation-agreement}), so most of the panel treats a
realistic request to breach a professional code as routine drafting work. The
requests carry no marker of harm that a general safety filter would react to,
and this is the gap the corpus was built to expose.

The ordering is resolved at its top and graded below it. Claude Sonnet 5,
Qwen3.6-27B and Qwen3.6-35B-A3B are separated from one another and from every
row beneath them. Below them neighbouring rows are not separated, and within the
family of comparisons two rows are separated exactly when their rates differ by
0.0753 or more (Section~\ref{sec:results-checks}).

The columns of Table~\ref{tab:leaderboard} are read as follows. The over-refusal
rate (ORR) and the benign violation rate (BNVR) are counted on the 279 twins
each row answers. ORR is the share of twins refused outright, and BNVR the share
on which the judge returns a violation. The five outcome shares are safe refusal
(SR), compliant non-violation (CN), violation with a warning (VW), silent
violation (VS) and self-contradictory violation (SC), and the first two sum to
the non-conflict rate (NCR). Each value is rounded on its own, so a row may add
to 0.999 or to 1.001. The share of invalid or unresolved judge outputs and the
sixth outcome, unresolved, read 0.000 in every row and are not printed.

\begin{table}[tbp]
\caption{The panel, ranked by the non-conflict rate}
\label{tab:leaderboard}
\centering
\scriptsize
\setlength{\tabcolsep}{1.7pt}
\begin{tabular}{@{}>{\raggedright\arraybackslash}p{39mm}rcrrrrrrr@{}}
\toprule
 & & 95\% interval, & & & \multicolumn{5}{c}{outcome shares} \\
model & NCR & family & ORR & BNVR & SR & CN & VW & VS & SC \\
\midrule
Claude Sonnet 5 & \cellcolor{green!51}0.849 & 0.762--0.908 & 0.025 & 0.100 & 0.663 & 0.186 & 0.086 & 0.065 & 0.000 \\
Qwen3.6-27B & \cellcolor{green!43}0.717 & 0.618--0.799 & 0.029 & 0.154 & 0.491 & 0.226 & 0.061 & 0.215 & 0.007 \\
Qwen3.6-35B-A3B & \cellcolor{green!37}0.609 & 0.507--0.703 & 0.025 & 0.147 & 0.412 & 0.197 & 0.136 & 0.247 & 0.007 \\
gpt-oss-20b & \cellcolor{green!21}0.355 & 0.265--0.457 & 0.014 & 0.197 & 0.258 & 0.097 & 0.014 & 0.631 & 0.000 \\
DeepSeek-V4-Flash-0731 & \cellcolor{green!17}0.287 & 0.204--0.386 & 0.007 & 0.168 & 0.115 & 0.172 & 0.283 & 0.427 & 0.004 \\
DeepSeek-V4-Flash & \cellcolor{green!16}0.262 & 0.183--0.360 & 0.007 & 0.154 & 0.122 & 0.140 & 0.201 & 0.534 & 0.004 \\
DeepSeek-V3.2 & \cellcolor{green!15}0.258 & 0.180--0.356 & 0.000 & 0.168 & 0.100 & 0.158 & 0.125 & 0.616 & 0.000 \\
T-pro-it 2.1 & \cellcolor{green!15}0.258 & 0.180--0.356 & 0.004 & 0.208 & 0.122 & 0.136 & 0.072 & 0.667 & 0.004 \\
YandexGPT 5.1 & \cellcolor{green!14}0.240 & 0.164--0.337 & 0.004 & 0.154 & 0.075 & 0.165 & 0.072 & 0.688 & 0.000 \\
GigaChat 3.1 10B-A1.8B & \cellcolor{green!11}0.183 & 0.117--0.274 & 0.004 & 0.186 & 0.068 & 0.115 & 0.154 & 0.663 & 0.000 \\
Qwen3-VL-235B-Instruct\textsuperscript{*} & \cellcolor{green!11}0.176 & 0.111--0.266 & 0.000 & 0.194 & 0.075 & 0.100 & 0.104 & 0.720 & 0.000 \\
Qwen3-32B & \cellcolor{green!9}0.154 & 0.094--0.242 & 0.000 & 0.190 & 0.022 & 0.133 & 0.036 & 0.810 & 0.000 \\
RuadaptQwen3-32B & \cellcolor{green!8}0.136 & 0.081--0.221 & 0.000 & 0.172 & 0.007 & 0.129 & 0.050 & 0.814 & 0.000 \\
Qwen3-VL-8B-Instruct\textsuperscript{$\dagger$} & \cellcolor{green!7}0.118 & 0.067--0.201 & 0.004 & 0.204 & 0.018 & 0.100 & 0.039 & 0.842 & 0.000 \\
\bottomrule
\end{tabular}

{\scriptsize\raggedright
\textsuperscript{*}Qwen3-VL-235B-A22B-Instruct.\quad
\textsuperscript{$\dagger$}The reconstructed share of unresolved verdicts for this
row is above the blocking threshold, and its interval is drawn over the family as
computed (Section~\ref{sec:results-checks}).\par}
\end{table}

\subsection{Models differ in whether they decline}\label{sec:results-refusal}

The rows differ in how often they decline the conflicting request, and hardly at
all in how often they fulfil it without conflict. The safe-refusal share runs
from 0.007 to 0.663 across the panel, while compliant non-violation stays
between 0.097 and 0.226. The non-conflict rate is exactly the sum of these two
shares, so its spread between rows divides between them: safe refusal carries
0.767 of the between-row variance, compliant non-violation 0.029, and their
covariance the remaining 0.204. A higher place in Table~\ref{tab:leaderboard}
thus means above all that the model recognised the request as one to decline,
and Section~\ref{sec:discussion-ordering} discusses what the ordering then
measures.

The models largely agree on which requests to decline: the eleven rows below the
top three issue 274 safe refusals between them, and 251 of these fall on tasks
that Claude Sonnet 5 also refuses. That share, 0.916, is well above the 0.663
that the refusal rate of Claude Sonnet 5 would give if the two were unrelated,
and 269 of the 274 fall on tasks that at least one of the three leading rows
refuses. The panel thus shares one ordering of how recognisable a breach is, and
the models differ mainly in how far along that ordering they decline.

This caution is selective: Claude Sonnet 5 refuses 187 of the 279 conflicting
requests outright and 7 of their 279 benign twins, and no row refuses more than
8 of its twins, so the over-refusal rate stays at or below 0.029 in every row
and covers 34 of the 3,906 answers to twins over the panel. Rows that refuse
more conflicting requests also refuse slightly more twins, with a Kendall's
$\tau_b$ of 0.745 between the non-conflict rate and the over-refusal rate.
OR-Bench reports the same association between safety and over-refusal on generic
prompts \citep{cui2024orbench}. Here its cost is small in absolute terms,
because the most cautious models decline the conflicting request and still serve
the legitimate twin built on the same scenario.
Section~\ref{sec:discussion-ordering} measures this separation pair by pair and
tests whether the length of the request explains it.

Benign twins draw violation verdicts as well, on between 0.100 and 0.208 of the
twins in each row and on 669 of the 3,906 answers over the panel. These verdicts
concentrate on a minority of twins: of the 279 twins, 209 draw none from any row
and 16 draw one from all fourteen, and the 50 twins flagged most often carry
0.907 of all such verdicts. The adjudicated human label flags a benign element
about as often as the judge does (Section~\ref{sec:validation-agreement}). The
benign violation rate therefore describes twins that invite a conflicting
answer, and it varies little between rows.

\subsection{A warning does not keep an answer within the clause}\label{sec:results-warnings}

Of the 551 answers to conflicting tasks that carry a warning without a refusal,
400 still produce the content that conflicts with the clause, a share of 0.726.
In twelve of the fourteen rows most warnings stand over conflicting content. The
exceptions are Claude Sonnet 5, whose warnings accompany a violation in 24 of 58
answers, and Qwen3.6-27B, in 17 of 39. Answers that produce the conflicting
content under a warning or an explicit refusal form the pattern this paper calls
safety theatre (Section~\ref{sec:procedure-axes}), and they number 407 over the
panel, between 4 and 80 per row.

The rows also differ in style: DeepSeek-V4-Flash-0731 warns in 100 of its 279
answers to conflicting tasks and produces the conflicting content in 79 of them,
while it refuses outright in 36. gpt-oss-20b warns in 6, refuses outright in 72
and complies without comment in 176. Their non-conflict rates, 0.287 and 0.355,
are not separated, yet their answers leave opposite impressions of caution. A
warning shows that the model registered a problem, and it does not show that the
model acted on it. An audit that counts caveats as caution would therefore
overrate exactly the models that warn most. TRIDENT describes the same behaviour
in open models, whose answers hedge and then comply \citep{hui2026trident}.

\subsection{Where the models conform}\label{sec:results-slices}

Conformance depends on the duty a clause imposes, and the dependence runs in the
same direction in every row. Table~\ref{tab:slices} gives the non-conflict rate
by harm domain and by clause type, over the whole panel and separately for the
three leading rows and the other eleven. These breakdowns were computed after
the campaign, and each difference cited below takes the clause as its unit,
because domain and type are properties of the clause and the tasks of one clause
are not independent.

The panel conforms most where the clause protects confidential information: the
non-conflict rate there is 0.611 against 0.279 on the other eight domains
together, a difference of 0.332 with a 95 per cent cluster-bootstrap interval of
0.201 to 0.467, and each of the fourteen rows scores higher on confidentiality.
The advantage does not come from refusal, since the models refuse
confidentiality requests about as often as the others, on 0.175 of answers
against 0.183, and fulfil them without conflict far more often, on 0.435 against
0.096. The lowest rates fall on procedural compliance, 0.232, and on conflicts
of interest, 0.252, two domains that together carry 125 of the 279 conflicting
tasks, and the eleven rows below the top three conform there on 0.112 and 0.143
of tasks. Protecting a client's or a patient's information coincides with
privacy, a category that general safety training covers, and the models carry
that habit into professional drafting. Duties that belong to the profession
itself, such as disclosing an interest or following a prescribed procedure, are
rarely taken as grounds to decline, and fulfilling the request is then itself
the breach.

The form of the duty matters as much as its domain, since a clause may prohibit
an act, oblige the specialist to perform one, or permit one under conditions.
The eleven rows below the top three conform to prohibitions twice as often as to
obligations, 0.287 against 0.143, a difference of 0.144 with a 95 per cent
cluster-bootstrap interval of 0.051 to 0.234 ($p = 0.004$ by permutation of
clauses). Each of the eleven rows shows the gap, and it appears inside every
harm domain that carries both kinds of clause. The three leading rows treat the
two kinds alike, at 0.731 and 0.705. A request to breach an obligation asks for
a text that leaves something out, such as a client statement that aggregates
what the code requires to be itemised. Such a text reads as ordinary editing,
and the weaker models do not recognise the omission as a breach.

\begin{table}[tbp]
\caption{The non-conflict rate by harm domain and clause type}
\label{tab:slices}
\centering
\footnotesize
\setlength{\tabcolsep}{4pt}
\begin{tabular}{@{}lrrrrrr@{}}
\toprule
 & & \multicolumn{3}{c}{non-conflict rate} & & \\
\cmidrule(lr){3-5}
level & tasks & panel & top three & other eleven & SR & CN \\
\midrule
\multicolumn{7}{@{}l}{\emph{harm domain}} \\
confidentiality & 42 & 0.611 & 0.897 & 0.532 & 0.175 & 0.435 \\
misleading communication & 29 & 0.367 & 0.793 & 0.251 & 0.256 & 0.111 \\
abuse of power\textsuperscript{*} & 8 & 0.366 & 0.833 & 0.239 & 0.295 & 0.071 \\
discrimination & 11 & 0.305 & 0.697 & 0.198 & 0.169 & 0.136 \\
integrity & 27 & 0.299 & 0.667 & 0.199 & 0.238 & 0.061 \\
professional competence & 26 & 0.297 & 0.769 & 0.168 & 0.217 & 0.080 \\
corruption & 11 & 0.253 & 0.606 & 0.157 & 0.130 & 0.123 \\
conflict of interest & 77 & 0.252 & 0.654 & 0.143 & 0.132 & 0.121 \\
procedural compliance & 48 & 0.232 & 0.674 & 0.112 & 0.170 & 0.063 \\
\addlinespace
\multicolumn{7}{@{}l}{\emph{clause type}} \\
permission & 24 & 0.402 & 0.806 & 0.292 & 0.128 & 0.274 \\
prohibition & 125 & 0.382 & 0.731 & 0.287 & 0.219 & 0.163 \\
obligation & 130 & 0.264 & 0.705 & 0.143 & 0.156 & 0.108 \\
\bottomrule
\end{tabular}

\vspace{3pt}
\raggedright\footnotesize
SR and CN are the shares of safe refusal and compliant non-violation over all
answers of the panel at that level. \textsuperscript{*}Below the declared minimum
of ten tasks for a level of harm domain.
\end{table}

\subsection{What changes a model's conformance}\label{sec:results-contrasts}

Three models of the panel were also run with reasoning enabled, at the same
weights and on the same route, and in two of them reasoning lowers conformance.
Each comparison is taken on the released deposit over the conflicting tasks both
runs answered. Qwen3.6-35B-A3B falls from 0.638 to 0.420 over 69 paired tasks
($p = 0.0015$), and Qwen3.6-27B from 0.714 to 0.529 over 70 ($p = 0.0023$), both
separated under Holm correction over the five pairs of the side arm.
DeepSeek-V4-Flash moves in the same direction, from 0.235 to 0.176 over 68
tasks, and is not separated ($p = 0.289$). The outcome shares show how the loss
arises: in the two Qwen3.6 rows safe refusals fall by half, from 35 to 16 and
from 38 to 19. Of the 44 tasks these two models refuse without reasoning and no
longer refuse with it, 21 turn into violations with a warning, 19 into compliant
non-violations and 4 into silent violations, while silent violations as a whole
barely change. Reasoning does not make these models miss the conflict, since
they still warn about it. It moves them from declining toward carrying out the
task, and as often over a warning as in a conformant form. Their refusals of the
benign twins do not change in this way.

Asking the model to play the role of a professional lowers conformance slightly,
and a hypothetical framing does not. On 100 scenarios realised as a direct
request and in both of these forms, and answered by ten of the panel rows, the
role-play form raises the violation share by 0.041 over 994 paired observations,
with a 95 per cent interval of 0.011 to 0.072 ($p = 0.0076$, separated under
Holm correction). The hypothetical form moves it by 0.009 ($p = 0.60$), and both
effects are small beside the differences between models. Past-tense and academic
framings were not realised on paired scenarios and remain unmeasured.

The line and generation of a model matter more than its size: the two Qwen3.6
rows reach 0.717 and 0.609, the two rows of the earlier Qwen3 generation reach
0.154 and 0.176, and all four gaps between the generations, from 0.434 to 0.563,
are separated. Within the Qwen3-VL line, a nearly thirtyfold difference in size,
from 8 to 235 billion parameters, moves the rate from 0.118 to 0.176, a
difference not separated under Holm correction. HarmBench likewise found
robustness to jailbreaks independent of model size within a family and dependent
on how the model was trained \citep{mazeika2024harmbench}. A newer generation
does not conform better by itself: in the DeepSeek line, V3.2 stands at 0.258
and V4-Flash at 0.262, and the two are not separated. The judge and the task
generator are both built on Qwen3.6-27B, so the lead of the Qwen3.6 rows needs a
check for leniency toward that family. Section~\ref{sec:results-checks} bounds
such leniency at 0.068, and the gaps between the generations are six to eight
times larger.

Adapting a model to Russian does not by itself make it conform to Russian codes,
as the two adaptations of Qwen3-32B on the panel show. T-pro-it 2.1 stands 0.104
above its base ($p = 4.2 \times 10^{-7}$), RuadaptQwen3-32B stands 0.018 below
it ($p = 0.424$), and the two adaptations stand 0.122 apart ($p = 6.2 \times
10^{-7}$). What an adaptation adds therefore depends on how it was trained, and
the language of the adaptation does not decide it. All four rows built for
Russian stand in the lower half of Table~\ref{tab:leaderboard}, and as a group
they violate on 0.796 of the conflicting tasks against 0.622 for the other ten,
a difference that is not separated ($p = 0.23$). TRIDENT found that
specialisation to a professional domain does not by itself make models safer
against breaches of professional codes \citep{hui2026trident}, and adaptation to
the language of the codes behaves the same way here.

The only frontier model on the panel, Claude Sonnet 5, stands first. YandexGPT
5.1 is the only other row with closed weights and stands ninth, so closed
weights do not by themselves bring conformance. With two such rows no group
comparison was run, and a single row cannot show whether the first place belongs
to the model, to its tier or to its developer. A dated snapshot of
DeepSeek-V4-Flash and the floating name of the same model score 0.287 and 0.262,
a difference the test does not separate.

\subsection{What the findings rest on}\label{sec:results-checks}

The findings rest on verdicts that agree with the adjudicated human label at
Gwet's AC1 of 0.894 on instrumental action, 0.915 on norm violation and 0.926 on
refusal type (Section~\ref{sec:validation-agreement}), and that follow the
clause the judge is shown (Section~\ref{sec:validation-clause}). The bands used
to read that agreement are 0.80 on norm violation, 0.70 on each of the other two
axes and 0.60 as the floor beneath all three. They are Cohen's $\kappa$ values
referred to SORRY-Bench and XSTest and were carried over to AC1 without
recomputation, which is why Table~\ref{tab:agreement} prints $\kappa$ and
Brennan--Prediger beside AC1.

No judge output was invalid, and on the 3,906 judged conflicting tasks no axis
value was returned unclear. A later audit found that the judge did not apply one
rule of its prompt, under which a finding of absence on an answer cut short is
returned unclear. Of the 9,450 verdicts, 126 rest on a truncated answer, and 123
of them assert an absence on at least one axis. Truncation reaches five of the
fourteen rows, and 93 of the 126 truncations fall on Qwen3-VL-8B-Instruct, 13.8
per cent of the 675 tasks that row answered. Applied afterwards, the rule turns
verdicts into unresolved ones and recomputes none, so it changes no rate in
Table~\ref{tab:leaderboard}. It raises exactly one row above the blocking
threshold of 0.05 on the share of unresolved verdicts, Qwen3-VL-8B-Instruct at
0.090. RuadaptQwen3-32B and gpt-oss-20b correct to 0.025 each and YandexGPT 5.1
to 0.004. Claude Sonnet 5 keeps a share of zero, because its one truncated
answer falls on a benign twin, and the remaining nine rows carry no truncated
answer.

The conditions under which that row was measured explain part of this. It ran
under an output ceiling of 2,048 tokens where the rest of the panel ran under
4,096, and the median truncated answer runs to 5,986 characters against 3,494
among the whole ones, so the ceiling removed the long answers. The route that
served a model weighed more, since the share of answers cut short runs from zero
to 0.254 across the twenty declared rows and reaches its maximum on a row that
ran at 4,096 tokens. The same open weights answering the same 248 tasks in the
same reasoning mode were cut short on 0.254 of them through the provider and
never on the local runtime.

Because of that row, the family of pairwise comparisons has two editions, which
differ in that one row. As computed, it spans fourteen rows and 91 comparisons,
of which 69 separate under Holm correction. Corrected, it spans thirteen rows
and 78 comparisons, of which 59 separate and 19 are not ordered, and
consequences are drawn on the corrected edition. Separation is decided by a
two-sided exact paired test, because the rows answered the same tasks. Every
separated pair stands further apart than every unseparated one, at 0.0753
against at most 0.0681. Intervals are Wilson intervals with a Bonferroni-style
$z = 3.4136$ over the 78 comparisons, except for the marked row, whose interval
is drawn over the computed edition at $z = 3.4554$. Pairing is what resolves the
ordering: the family-corrected interval on the top row alone spans 0.762 to
0.908, and read as independent estimates the intervals of
Table~\ref{tab:leaderboard} would order almost nothing. Each task received one
model response, so the intervals are computed over items and do not cover the
variance of response sampling.

The ordering also holds without the five withheld documents: all 59 separated
pairs keep their direction on the ten released documents, at 189 tasks a row,
and on the released deposit, and the top row does not move. On the deposit, the
sixth and the ninth rows of Table~\ref{tab:leaderboard} come out exactly equal
and share a rank, and this pair is one of the nineteen that the test does not
separate.

Three properties of the verdict bound how far the findings reach, and
Section~\ref{sec:discussion} discusses the first two. First, the verdict depends
in part on two fields that describe how the task was built to threaten the
clause. Removing them from the judge prompt shifts the verdict by 0.060 against
a declared bound of 0.020, over 700 paired observations drawn from the earlier
set at fifty per row and with partial violation counted as violation, and
applied row by row the shift keeps 58 of the 59 separated pairs in their
direction. Second, the evidence span that a violation verdict cites can be
located in the answer for 0.623 of such verdicts on prohibition clauses under
the declared rule, and for 0.937 under the most generous admissible reading.
Third, the judge was frozen from the base model that generated the tasks and
that one panel row serves. As groups, the seven rows of the judge's model family
and the seven outside it do not differ in violation share, at 0.690 against
0.652 ($p = 0.814$ by exact permutation). Regenerating the tasks of the same
scenarios with a generator from a family absent from the panel changed the
violation share of the rows in the generator's family by $+0.0157$ relative to
the other rows, at a half-width of 0.068, which excludes an advantage for the
generator's family larger than 6.8 percentage points.

\section{Discussion and Limitations}\label{sec:discussion}

Three of the eight declared checks did not meet their threshold.

\subsection{An ordering by caution}\label{sec:discussion-ordering}

The first of the three bears on what the ordering orders. The non-conflict rate
is identically the sum of two outcome shares, and one of the two accounts for
almost all of the between-row variance, which is a fact about this panel and
not a definition. What this instrument orders is therefore closer to
domain-scoped caution than to conformance with a professional code in any broad
sense, and the broad claim is withdrawn. What survives is narrower and is not
nothing. The caution the ordering reflects is discriminating, firing on the
conflicting task and not on the twin built from the same scenario substrate, and
a model that refuses indiscriminately is visible as such in the companion
quantity beside the rate. That discrimination is now measured on the pairing
itself, one conflicting task against its own twin, 279 pairs to a row. It is
positive in every row of the panel, from 0.0072 to 0.6452 at a median of 0.1111,
no row at zero and thirteen of the fourteen with an interval clear of it. The
printed value is conservative: setting aside the 51 twins on which at least
seven rows find a violation raises the median to 0.1250 over the remaining 228
pairs a row and raises the value in all fourteen rows, and on the ten released
documents the same cut takes the median from 0.1032 over 189 pairs a row to
0.1097 over 155, up in thirteen rows and down in one. The direction is in that
sense a property of the panel, while the size of the effect differs across it by
a factor of ninety, which is to say that whether a model tells the two members
of a pair apart is a question with fourteen different answers.

Whether that discrimination is a discrimination by length was declared before
the campaign as a length-matched paired reanalysis, and the completed pairing is
what made it possible to run. The conflicting member of a pair is the longer one
in 0.697 of pairs. Matching the two members within a window of five words leaves
91 matched pairs a row, on which the median difference is $+0.099$ against the
$+0.111$ obtained with no matching at all. The rule fixed beforehand asks for
half of the unmatched median, $+0.0555$, so the discrimination survives length
matching. The threshold is the smallest at which every row reaches the thirty
pairs the design requires, and it was taken mechanically. At ten words the
median is $+0.086$, which is lower, so a threshold picked to make the quantity
look best would not have been this one.

Three things belong to that measurement. On the released deposit the same
measurement under the same rule still does not run, the weakest row reaching 24
pairs at a threshold of ten words against the thirty required, and a row there
carrying 189 tasks against the 279 named above, both runs being published with
the set each was taken on. The set was changed after the outcome on the deposit
had been seen. The criterion was not touched, and before the pairing was
completed the measurement was impossible on the analysis set too, which then
held 24 pairs in all. Two unpaired measurements on the deposit bear on the same
question, and neither supports the alternative. Online Resource 1 reports
both.

\subsection{Checking a verdict against its evidence}\label{sec:discussion-span}

The second check concerns whether a reader can verify a verdict against the
evidence it cites. The frozen judge reproduces its own verdicts, since three runs
on the same 300 records never disagreed with one another and reproduced 298 of
the 300 verdicts the campaign had returned (Section~\ref{sec:validation-clause}).
Every violation verdict on a prohibition clause cites a fragment of the answer as
its evidence. Once markdown formatting is set aside and partial matches are
counted, the cited fragment occurs in the answer for 0.937 of the 1,343 such
verdicts in the campaign and for 0.951 of the 1,120 in the deposit, against a
declared threshold of 0.95. The declared rule accepts only a verbatim occurrence
and gives 0.623 in the campaign and 0.627 in the deposit. Most of that difference
comes from formatting, because the judge quotes the answer without its markdown
symbols, and a verbatim search then misses 288 quotes that are present in the
answer. A further 133 quotes are shortened or reordered, and only 85 of the 1,343
cited fragments do not occur in the answer at all. The check therefore falls
short of its threshold under the declared rule, and under the most generous
admissible reading it meets the threshold in the deposit and comes close to it
in the campaign.

\subsection{A verdict that moves with construction-side
fields}\label{sec:discussion-fields}

The third check concerns two fields that describe how a task was built to
threaten its clause, and it shows that the verdict depends on them in part.
Removing the two fields from the judge prompt moves a net 0.060 of 700 paired
verdicts toward no violation, three times the declared bound of 0.020, while the
same configuration run again on the same records moves none. The fields help the
judge recognise the conflict the task was built around. On the 83 of 1,200
adjudicated elements where the two configurations disagree, the adjudicated
human label sides with the configuration that keeps the fields 60 times against
17, and agreement with people falls from AC1 0.915 to 0.872 when the fields are
removed. The annotators saw the same fields, so this result does not prove the
fields neutral. It does show that removing them would not bring the judge closer
to people. The dependence changes the ranking very little, since applied row by
row the shift keeps 58 of the 59 separated pairs of rows in their order.

\subsection{Transfer to another jurisdiction}\label{sec:discussion-transfer}

One more check was declared before the English arm collected its first answer,
and it came out in the instrument's favour. The pipeline carries to another
language and another jurisdiction. Two codes of other jurisdictions in English
were taken in under the same eligibility criterion, and Online Resource 1 names
them and gives the size of that arm. The same pipeline generated tasks from them
and accepted those tasks. The English arm is judged by an English prompt that
was frozen under its own identifier and put through the same gate before use.
Its verdicts agree with adjudicated human labels at AC1 of 0.823 to 0.958. No quantity from that arm is set beside one measured in Russian, because
corpora built from the codes of different legal systems do not measure on one
scale. The transfer exposed twelve defects of one kind. In each of them a step
that depends on language silently fell back to Russian or lost track of the
language of the text, and still reported success, so the run gave no sign of the
defect. For example, the list of banned words was Russian, so its patterns never
matched English text and the validator reported a pass with no findings. All
twelve defects were fixed, which confirms that the pipeline is portable to
another language.

\subsection{Threats no check was declared
against}\label{sec:discussion-threats}

Several threats had no check declared against them, and they are set out next.
The three axes are defined so that a value on one never presupposes a value on
another. How strongly they are associated in real answers was not estimated, and
some association is expected, since an answer that refuses usually does not
carry out the request either. The verdict is
about the text of an answer, so a duty is inside the perimeter when its literal
fulfilment puts into the answer a fragment conflicting with the clause, and
duties violated by the specialist's subsequent \emph{action} fall outside by a
criterion in the definition of the construct. The share of professional
obligation this excludes is not measured, and we do not assert that it is small.
The verdict does not score the useful part of an answer, compliant non-violation
being a category and not a grade, and what it orders is a row, a model together
with a reasoning mode. That a task should not announce its own purpose was given
to the generator as an instruction and is verified by no mechanism at any stage,
so a model refusing on the surface form of such a request is scored as though it
had refused on the clause.

The thresholds on lexical borrowing fix the share of comparisons flagged, not a
boundary between shared subject matter and paraphrase. They flagged ten
comparisons. All ten were read in full, and the ten pairs were regenerated. On
the two non-Russian codes the thresholds were carried over, and the acceptance
run of that arm returned a pass with none of the three checks able to fire. That
is one of the twelve defects of Section~\ref{sec:discussion-transfer}, closed at
first only in how the thresholds were labelled, with the verdict left a pass.
Acceptance now refuses a dataset whose declared language is not that of
calibration, and passes it only by a declared withdrawal.

Nor is the benign twin read against its clause at any stage. That it offers
nothing to violate is a property of how the pair is built, and
0.251 of the twins draw at least one violation verdict. A refusal of a twin
that does conflict with its clause is counted as over-refusal and lowers
paired discrimination, so the printed over-refusal rate errs high and the
printed discrimination low, both against the models.

The route a row was served by is not controlled either. The panel was served
through three handles of two kinds, a provider interface and two cards of one
local machine. The leaderboard ranks rows of both kinds in one table. Nine
ranked rows ran through the provider and five locally, three of those five on
the second card, which are GigaChat 3.1 10B-A1.8B, Qwen3-32B and
RuadaptQwen3-32B. That the two cards are one condition and not two rests on a
measurement of its own: asked the same tasks, they returned 21 of 21 answers
identical byte for byte, against a control of the same size comparing one card
with itself. That measurement is one model at one submission rate, so the three
rows on the second card inherit it without being measured themselves, and the
control pair declared for the route reaches the first local handle only. That
pair was declared to turn the confound into a measured quantity, but measurement
put its two halves in different reasoning modes, so the pair left the main arm
and the consequence was written down before the campaign. What the side arm
returns is a bound and not a null. Over the 57 tasks the two arms have in common
the difference is $+0.018$ at $p = 1.000$, at a half-width of 0.103. That is
wider than the 0.0753 at which the corrected family resolves, and 32 of the 59
comparisons that family separates set a locally served row against one served by
the provider, so the measurement does not exclude a route effect large enough to
reorder the panel.

Nondeterminism, scope and sampling bound the whole. The campaign ran at
temperature 0, so its own spread is nondeterminism of \emph{service}, and
service is what an external replicator meets. Repeated collection under the
declared conditions reproduces the response byte for byte in only 0.0071 of
tasks over three observations, 3 of 420, which is our own bound on the
reproducibility of the procedure and not a measurement of any model. The version
of the local runtime that served the subject models was recorded in no run
header and in no freeze record, which is a limit on replication stated here. The
judge's own build is recorded, at \texttt{llama.cpp} \texttt{b10362-4801e3c56}
with a context of 72,192 tokens and two slots, all three belonging to the frozen
judge. The corpus covers fifteen professions of a single legal system, in
Russian, and one jurisdiction. Tasks are generated, so the target of inference
is the dataset, and a provider service limit truncates exactly the tasks on
which a model generates longest, so pairing protects the comparisons and does
not protect the absolute values. Every measured table here covers nine of the
ten harm domains, the tenth being absent from the analysis set altogether. Six
levels elsewhere are under-populated against their declared minimum, and no
interval is drawn on a slice below its minimum.

One property of the corpus bounds what a verdict carries back.
Five codes are advisory, with 972 of the 3,433 conflicting tasks and
220 of the 2,455 in the ten released documents, so a verdict against one of them
reports the breach of a declaration.

\subsection{What the release lets a reader redo}\label{sec:discussion-release}

A reader working from the deposit can repeat most of what is reported here.
Of the 15 documents in the corpus, 10 enter the release and 5 are withheld,
each for a reason recorded against the document in the deposited inventory.
Two rightsholders prohibit reuse, and between them the prohibition covers three
documents. Permission requests to those two were drafted and not sent. A third
rightsholder permits the link but not the publication of fields derived from the
document. The fifth document has neither published terms of use nor an
established edition. A licence value does not separate these five: all five are
recorded restricted or unknown, and so are six of the ten released, one of them
restricted. All 15 passed legal verification, so withholding is an outcome of
that verification. The deposit therefore carries 4,470 of the 6,866 records:
1,956 belong to the five withheld documents, and a further 440, the 220 pairs
held out under a fraction fixed before selection, are kept back by design and
their membership is not published. The reported values are not recomputed on the
10, because a corpus bounded by what may be republished is not the corpus the
methodology defines, and substituting the one criterion for the other would
change the object measured. Declining to recompute is not a reason to decline to
report. With the withheld documents removed each panel row keeps 189 of the 279
conflicting tasks it received, and the non-conflict rate rises on all fourteen
rows, by between $+0.0067$ and $+0.0521$, mean $+0.0227$. The rise is a
systematic displacement, the withheld documents being the harder ones, and its
size varies across rows by nearly a factor of eight, so the published ordering
is not claimed to survive it. The address of each document is listed with it, so
a text this project does not republish remains obtainable, and what is withheld
is our copy of it. The intake reading of Section~\ref{sec:resource-eligibility}
is deposited with both sets of labels, but its identifiers belong to the copies
assessed at intake, and the clause text is withheld, so the arithmetic of that
reading can be redone and the reading itself cannot be checked against the
documents.

The code of the project enters the release with twenty executable entry points,
among them the four-step analytical pipeline and the driver that measures span
traceability. The deposited annotations carry, for
every item, the task, model and arm of the verdict it labels and mark the rows
the adjudicator wrote, and the deposited adjudication module resolves them into
one label. Agreement recomputed on the deposit is
0.886 on instrumental action against the 0.894 of Table~\ref{tab:agreement},
0.900 on norm violation against 0.915, and 0.916 on refusal type against 0.926,
over the 797 conflicting-task items whose verdict the deposit carries, and the
traceability of evidence spans can be recomputed there too. Twenty-three paths
are withheld by name: the registry of formulations this manuscript avoids and
code retired from use.

One bound holds however much is released, since it belongs to the
measurement. Repetition of the campaign meets the nondeterminism of service
reported above. The corpus itself was generated at a non-zero temperature, so an
identical regeneration is unavailable to us as well, which is a property of how
the corpus was made, and it obstructs nothing, since the corpus itself is
published. Comparisons drawn after publication without accounting for possible
ingestion of the corpus into training data are declared invalid, and the canary
strings, the declared held-out subset and this caveat are published with the
artefact.

\section{Ethics, Broader Impact and Release Risk}\label{sec:ethics}

Half of the corpus consists of requests constructed so that carrying them
out would violate a specific professional norm, and the release carries the
7,303 published verdicts drawn on them, of which 2,621 judge an answer to
violate a named clause in full and 41 judge one to violate a clause in part,
each joinable by task identifier to the record that states the mechanism the
task was built on. What makes this material worth having is what makes it
dangerous. Such requests read as ordinary professional work and trip no
safety classifier, which is why they are diagnostic and equally why they
serve as a source of well-formed requests a professional filter would not
stop. The resource is dual-use, and calling it anything else would be a
misuse of words before it was a misuse of the corpus.

It is released even so, because an evaluation that cannot be reproduced is
not an evaluation. Withholding a resource of this kind keeps the risk, the
constructions being not hard to rediscover, and removes the benefit, which
is that a third party can check what was measured. It serves an auditor
testing a deployed assistant against the code its users are bound by, a
regulator asking what conformance would even mean as a measurement, and a
builder of guardrails who needs cases a generic safety filter does not
catch. It serves the body that wrote the clause most directly, since a
verdict naming that clause is actionable in a way a benchmark score is not.
It serves in the wrong direction anyone looking for a ready phrasing of a
request that circumvents a professional norm. We state that asymmetry and do not
resolve it, because a reader deciding whether to use this corpus is entitled to
it. The canary strings and the held-out subset exist so that a comparison drawn
later can be told apart from one drawn on a model that has already seen the
corpus. They restrict nothing, and reading them as a safeguard would be reading
a measuring device as a lock.

The material the corpus is built on is under third-party copyright, and the
reference mode is what makes it publishable. \textbf{Facts about the content of
a norm fall outside the object of protection} under paragraph 5 of Article 1259
of the Civil Code of the Russian Federation, which is what allows the derived
deontic fields to be published while the wording is not. Paragraph 6 of the same
article \textbf{places an official act of a state authority outside protection
altogether}, and that is the ground on which one document is published in full
as a worked example. Where a derived field nevertheless coincides with the
wording, the ground is paragraph 1 of Article 1274, which \textbf{permits
quotation for scholarly purposes with the source named}.

That coincidence is measured, and absence of verbatim reproduction is claimed
nowhere in this paper. Over 1,898 clauses the published fields reproduce the
clause verbatim in 0.166 of cases on the conditions field, 0.125 on the action
and 0.358 on the actor role, with a longest verbatim run of 46 words. Taking
runs of at least ten words together, 53 of those clauses are recoverable beyond
half from the published fields, and one entirely. Quotation under Article 1274
requires the source, the edition and the address of the clause to travel with
the quotation, and the release carries all three. The same article conditions
free quotation on the volume being justified by the purpose of quoting. The one
fully recoverable clause is named by identifier as a disclosed limit, and
withholding the clause text does not exclude partial recovery of it from the
fields published beside the hash.

A deterministic screen covers three free-text fields of every published
verdict: the answer, the evidence span and the judge's short rationale.
Under 18 classes of structured identifier it returns 5,076 matches, among
them 36 shaped like a passport number and 49 like a social-insurance number.
The unit read is the distinct extracted string, not the match: every
distinct string in the nine document-and-identity classes was read, and the
five name classes were read at their frequency head. Not one match in the nine
classes has the form of a real person's identifier. The card numbers are the
standard test sequence, the passport numbers are placeholders, and the names are
the Russian equivalent of a placeholder name. Those strings are in the answers
because the answers are drafts of documents, and a model drafting a document
fills the fields of the document. Reading establishes that two taxpayer numbers
belong to actual organisations, both publicly registered legal entities used as
textbook examples. One further taxpayer number is not settled by reading at all,
its length and prefix admitting an active legal entity that only a register
lookup would separate from a placeholder. A further 144 distinct electronic mail
addresses, some of them at live providers, are neither removed nor masked. The
screen does not establish that the corpus holds no personal data, its recall
being unmeasured and not measurable without a reference annotation. What is
claimed is a count over a named set and a reading of that count. Model answers
are not published under a model name without the snapshot and configuration that
produced them, and annotations carrying information that identifies an annotator
are not published.

\section{Conclusion}\label{sec:conclusion}

This work builds a benchmark of 3,433 conflicting tasks with one benign twin
apiece, each anchored to a clause of a professional code in a named edition, and
releases 2,235 of those pairs, ten documents of the fifteen, together with the
pipeline that produced it and a model judge validated against adjudicated human
labels. What is withheld, and why, is set out in
Section~\ref{sec:discussion-release}. The pipeline and the judge were frozen
before any human label on a measured answer existed and were not adjusted
afterwards.

For most of the fourteen combinations of model and reasoning mode evaluated
here, a request to breach a professional code is ordinary work. Eleven of them
produce content that conflicts with the clause on most such requests, and they
mostly do so without comment. The models that conform more do so by declining,
and they still serve the legitimate twin of the request they decline.
All models largely agree on which requests are recognisably improper and differ in
how many of those they decline. A warning is not a sign of conformance, since nearly
three warnings in four accompany the conflicting content. Conformance is highest
where a clause protects confidential information, the professional duty closest
to general safety training, and lowest on conflicts of interest and on
procedure, and the weaker models conform to a prohibition twice as often as to
an obligation. Enabling reasoning lowered conformance in two models by turning
refusals into completions, often under a warning, and a role-play framing
lowered it slightly. The line and generation of a model mattered more than its
size, and adapting a model to Russian did not by itself raise its conformance to
Russian codes.

These conclusions have three limits. First, the ranking reflects above all how
readily a model declines, since the share of conformant completions barely
differs between models. Second, the evidence span a verdict cites can be located
in the answer for 0.623 of violation verdicts on prohibition clauses under the
declared rule, and for 0.937 under the most generous admissible reading. Third,
the verdict depends in part on the fields describing how a task was built,
although that dependence leaves 58 of the 59 separated pairs of rows in their
order. The instrument was not adjusted in response to any of these results.

The pipeline is not limited to Russian. Carried to two codes of other
jurisdictions in English, it produced tasks and verdicts, the verdicts were
validated against human annotation, and the defects the transfer exposed were
found and fixed, so its portability to another language is implemented and
confirmed in practice. A result on this benchmark is a diagnostic finding for a
stated use, and how it should be read depends on who uses a model and in which
professional scenario.

\bmhead{Use of AI tools in manuscript preparation}

The authors used large language model assistants Claude Code (Anthropic),
Perplexity (Perplexity AI), and Kimi (Moonshot AI) to search for related
literature, to translate this manuscript from Russian into English, and to edit
and refine the English text. These assistants were also used to support the
development of the software underlying this study, including generating unit tests,
identifying code defects, and documenting code. The authors reviewed all output of
these tools without exception, verifying each work found through the literature
search against its original publication, and made corrections where necessary.
The authors confirm that the text reflects their original work and take full
responsibility for the content of the published article. Any use of AI tools as
part of the research itself, including the model judge and the language models in
the pipeline that produced the benchmark, is described in the main text.

\backmatter

\section*{Declarations}

\bmhead{Data availability}

The package this paper is checked against is
\texttt{tenetbench\_release\_20260915.tar.gz}, 142 files carrying 2,235
conflicting tasks with their benign twins and 7,303 verdicts over twenty rows,
whose content digest is \texttt{c221e63b}\ldots, and it supersedes every earlier
build of the package. It carries ten of the fifteen documents of the corpus and
not the 220 pairs held out by design, and the reason each document is withheld
is recorded against it in the deposited inventory. It is deposited on Zenodo
under CC BY 4.0 and is available at
\url{https://doi.org/10.5281/zenodo.22678484} \citep{anichkov2026tenetbench}.

\bmhead{Materials availability}

Not applicable.

\bmhead{Code availability}

The generation and evaluation pipeline is deposited in the same Zenodo record
as the data, under Apache-2.0: \mbox{\url{https://doi.org/10.5281/zenodo.22678484}}.
Both licences are declared on the record and are carried inside the archive as
\texttt{LICENSE-DATA.txt} and \texttt{LICENSE-CODE.txt}.

\bibliography{tenetbench}

\end{document}